# Generalized Reinforcement Learning: Experience Particles, Action Operator, Reinforcement Field, Memory Association, and Decision Concepts

Po-Hsiang Chiu[1], Manfred Huber[1]

*Abstract*— Learning a control policy capable of adapting to time-varying and potentially evolving system dynamics has been a great challenge to the mainstream reinforcement learning (RL). Mainly, the ever-changing system properties would continuously affect how the RL agent interacts with the state space through its actions, which effectively (re-)introduces concept drifts to the underlying policy learning process. We postulated that higher adaptability for the control policy can be achieved by characterizing and representing actions with extra "degrees of freedom" and thereby, with greater flexibility, adjusts to variations from the action's "behavioral" outcomes, including how these actions get carried out in real time and the shift in the action set itself. Moreover, in standard RL, the inherently restricted definition of actions and the corresponding (action-induced) state-transition mechanism, tends to result in intractably large state space coupled with open questions on how the learned policy generalizes to the unknown part of the state space, a perpetual puzzle to solve in a time-varying environment. This paper proposes a Bayesian-flavored generalized RL framework by first establishing the notion of parametric action model to better cope with uncertainty and fluid action behaviors, followed by introducing the notion of reinforcement field as a physics-inspired construct established through "polarized experience particles" maintained in the RL agent's working memory. These particles effectively encode the agent's dynamic learning experience that evolves over time in a self-organizing way. Using the reinforcement field as a substrate, we will further generalize the policy search to incorporate high-level decision concepts by viewing the past memory as an implicit graph structure, in which the memory instances, or particles, are interconnected with their degrees of associability/similarity defined and quantified such that the "associative memory" principle can be consistently applied to establish and augment the learning agent's evolving world model.

*Index Terms* — parametric actions, kernel methods, metric learning, function approximation, Gaussian process regression, reinforcement learning, action operator, reinforcement field, associative memory, abstract actions, functional clustering, spectral clustering.

## I. Introduction

COMPLEX control domains, involving high-dimensional feature space and potentially time-varying and evolving systems, have been a long-standing challenge for reinforcement learning (RL) due to limitations in the state-space and the policy representation. The standard reinforcement learning framework [4] establishes a unique way of deriving control policies through the state abstraction of the environment that responses to a set of actions in a manner subject to the environment dynamics. A value function maps input states or state-action pairs to their corresponding utility estimates, which serve as a fundamental basis for evaluating a control policy. However, most of the existing applications of RL rely on a prior knowledge of action choices, such as the available actuators in robotic control domains, etc. The complexity of RL algorithms thus depends largely on the dimensionality of the state space abstraction, which then governs how effectively the utility of a particular decision can be evaluated. The challenge arises when applying RL methods to real-world control tasks that require numerous variables to be accurately represented, leading to an exponential growth in the state space and difficulty in representing the policy.

The barrier from the state space modeling is also rooted deep down in the modality of the policy learning mechanism and its representation. First, the abstraction of a given task is largely expressed through the features encoding the state space, which often requires maintaining a global system state. An extreme case can be observed from using multiagent approach to address decentralized MDP problems [22] such as applying the cooperative multiagent MDP framework in task assignment domains [23]. In such framework, user tasks and compute resources are represented by their individual state and action spaces, and as a result, learning a task assignment policy requires forming a joint state space, say, as the Cartesian product of the states associated with all machines and all tasks. The computational cost of maintaining a global system state as such can be exceptionally high if not intractable for a large Grid network [21]. Second, the control policy is expressed over a concrete set of actions, each of which effectively models a set behavior applicable under certain situations as characterized by related states; however, most policy learning methods do not fully utilize such *aggregate states*, as addressed by the factored-MPD framework [11], where only a consistent subset of actions takes effective steps towards the goal. Consequently, much computational effort is spent in the state space enumeration during the policy evaluation.

---

[1]The authors are with the Department of Computer Science and Engineering, University of Texas, Arlington, TX 76019 USA (e-mails: po-hsiang.chiu@mavs.uta.edu; huber@cse.uta.edu).

On the other hand, performing actions often involves certain behavioral details characterized by a dynamic process contingent on not only the variation from within the internal state of the agent but also from the continuous variation in the environment with which the agent interacts. Expressing such a dynamical process, referred to as *action dynamics* later, would require a set of parameters being controlled simultaneously and moreover, these parameters are often correlated. In automatic vehicle steering task [24], for instance, more than a few physical attributes are involved in specifying a particular choice of motion control (as an action) such as steering the vehicle within a certain range of velocity following the direction expressed in a vector form in which each vector component is constrained within a tolerable error range.

Furthermore, the standard RL framework often lacks the ability to efficiently adapt to an evolving environment or unexpected behaviors associated with the actions in that the agent is not aware of the "meaning" of actions beyond the policy being followed. As RL algorithms are typically developed in simulated environments, uncertainty inherent in action behaviors can also arise when porting the learned policy to physical devices (e.g., a physical robot executing a learned policy). Yet, these two seemly different issues, i.e., action dynamics and policy adaptability, are indeed closely related. Consider a robotic navigation domain, for instance, the varying environment dynamics could be related to a newly formed, bumpy surface en route or a mis-calibrated motor that occurs only after an optimal policy is learned. As a result, the agent may deviate away from the course where the optimal policy is feasible without adjusting the policy in parallel. Similarly, variations in the available action set can also alter the agent's ability to perceive and interact with the environment. Using the standard RL, performing real-time error recovery for skewed action outcomes or policy adjustments are not possible without re-learning the policy from the start since the varying environment is effectively redefining the underlying RL problem.

*A. RL Generalization I: Incorporating Action Dynamics*

To address the uncertainty in the ever-changing environment and fluid action behaviors (e.g., stochastic outcomes, variations in the permissible action set), perhaps we will need to start from shifting the perspective of the existing RL framework slightly. Our first objective is to cope with the uncertainty associated with time-varying action dynamics while considering compactness of the underlying representation because only then will the new RL framework scale to complex domains that would involve potentially a multitude of variables not only for the state but also for the action.

We will begin by considering three mainstream methods that aim to scale RL through achieving higher policy adaptability and a reduction in the representational size: i) value function approximation ii) policy generalization, and iii) state/action space abstractions. As will become clear in the following sections, these three directions can be unified into a consistent learning framework through more flexible, dynamic action representations coupled with an associative memory that systematically updates itself as the RL agent continues to accumulate its (policy) learning experience.

This new RL paradigm is referred to as the generalized reinforcement learning (GRL), which, at its core, attempts to link together policy search and decision concept formation while encouraging compact policy representation. Methodology-wise, we will focus on a kernel-based, Bayesian approach to realizing GRL in this paper, which is perhaps the most straightforward (but by no means the only) way to implement such systems. We will briefly examine why that is starting from the function approximation perspective.

Recent work in function approximation methods [2] have been used extensively to ameliorate the so-called curse of dimensionality in complex systems by using a selective set of basis functions to represent the value function. Challenges arise in choosing the right basis set that, if sufficiently large, generalizes well enough to reflect the topology of the state space. However, the size of the basis set can also grow rapidly with the dimension of the state space. Proto-reinforcement learning framework [3] identifies a task-independent basis set and its structure from the samples of prior learning experience by analyzing the state space manifold encoded via a graph Laplacian.

In the GRL formulation, we envision a system that not only supports value function approximation but, in parallel, also supports metric learning that allows for the RL agent to relate *decision contexts* (Section V.B) from one another in a "self-organizing" way: besides policy search, the agent also learns to establish a similarity metric that operationally plays multiple roles such as identifying, updating (when necessary), and grouping state-action pairs while factoring in their corresponding value estimates. To satisfy these desirable properties, Gaussian process regression (GPR) becomes a natural candidate to represent the (action-)value function since the training of GPR-based models is driven by samples of experience but without needing to learn the basis function explicitly. Meanwhile, the underlying "machinery" of the GRR, as a nonparametric kernel-based regression model, helps to further integrate other required policy-learning constructs, based upon the decision context-specific similarity metric, that morphs and adapts to the changing environment. Before moving forward, we note that related work in RL based on gaussian processes (GP) can be found in [5], [25], where the GP, in addition to value functions, has been used to represent the environment dynamics including the state transitions, observations, and reward functions, etc. More on the GP in the context of GRL will be given in Section II.D.

GRL extends upon the standard RL framework by first introducing extra degrees of freedom to the action set such that the local dynamics within the state transition and the policy search can be integrated as a unified continuous process. Specifically, actions are treated collectively as a parametrizable, operator-like construct that encodes dynamic processes underlying state transitions. We further introduce the notion of *action operator* that specifies how an action "operates" on its input state, alters its configuration stochastically, and then produces an output state (Section III.A). The action operator is linked to the value function through *action parameters* and properties of *kernel functions* [15], which enables systematic estimation of *fitness values* for any combinations of state and action. We use the term "fitness" because both the state and the action in GRL are defined more broadly as special entities that interact with each other and hence, the degree of fitness is used to quantify how well a state and an action "bond" with each other, much akin to the role of utility estimates in standard RL. The exact behavior of the action operator depends on its parameters as random variables with their values stochastically determined by the task-specific random processes. Parametric actions and action operator will be covered in Section II.B and III.A respectively.

Next, with actions being parameterized in a manner that exhibits properties of a continuous function, we further establish the notion of the *reinforcement field* as a policy generalization mechanism. The property of the "field" comes from using GP as a kernel machine, in which kernel functions play the role of adaptive correlation hypotheses over *experience particles*, which represent the decision-specific learning experience gathered over time as the RL agent continues to explore the environment and operates in the vector field of functions, or more specifically, the reproducing kernel Hilbert space (RKHS). We will dedicate an entire section for reinforcement field in Section III.

The value function in the reinforcement field is driven by a pre-specified temporal-difference (TD) learning algorithm that updates the RL agent's (inherently limited) working memory holding the experience particles that are deemed essential for (fitness) value predictions. Consequently, the policy evaluation and improvement also depend on the self-organizing sampling and updating processes of relevant memory instances (or particles). To this aim, a TD-guided MCMC sampling approach, referred to as *particle reinforcement* (Section IV), is used to update the agent's working memory in a manner that that encourages the learning agent to select actions leading to positive increments of fitness values (i.e., positive TDs) while, by symmetry, preserving memory instances that led to negative increments (i.e., negative TDs), serving as constraints for the agent to avoid repeatedly taking undesirable actions.

*B. RL Generalization II: Learning Decision Concepts*

Establishing a high-level abstraction over the state space and/or within the policy representation is yet another approach to scaling RL algorithms and increasing their adaptability to complex systems that are subject to change. Hierarchical reinforcement learning (HRL) based on clustering [19] and/or subgoal discovery [13] decomposes the otherwise large state space into different smaller regions where sequences of actions (i.e., the options formulation [26]) are determined for the associated subtasks, each of which constitutes an intermediate step necessary to reach the final goal (hence the name subgoal). With the HRL, only a subset of the state space subject to change would require updates on the learned policy. The complexity in the policy learning is also greatly reduced due to a reduction of the state space through learning the local policy in each subgoal area individually. Factored MDP, on the other hand, reduces the representational size by expressing states implicitly through a set of random variables [10]. Subsequently, a dynamic Bayesian network (DBN) is established as a compact transition model based the rationale that the transition of one variable often depends only on a relevant subset of the other variables. The abstraction is thus built upon the structural regularity within the state transitions in response to the stochastic effect of actions. Doing so effectively aggregates a set of states that responds to an action in a consistent manner. Additionally, the DBN representation can be further extended to capture the *context-specific structure* [10] from within the conditional probability table (CPT) that exhibits patterns in terms of the assignments to a subset of variables upon state transitions. That is, certain state-action pairs can only occur in relevant contexts in which the transition involving the relevant pair is a feasible choice (effectively with a non-zero probability).

In light of the hierarchical and context-dependent algorithmic constructs from the aforementioned methods, we will further establish a (decision) concept-driven learning architecture (CDLA) in GRL that enables the learning of levels of abstraction over similar decision contexts, a terminology used to emphasize over a set of "compound state-action pairs" and their equivalency in the role of policy generalization, where the state provides the context in which appropriate actions apply with fitness estimates as responses. Note that, based on the parametric-action model, the concept of state-action pair can now be generalized to broader sets of representations. By leveraging the property of kernel functions, for instance, state and action can, in principle, take on any forms (e.g., vector, string, graph) so long as their inner products are well-defined through the definition of valid kernel [15]. In this paper, we will focus on the vector representation.

Learning levels of (decision) context-specific abstraction can be realized through *kernelizing* the (generalized) state-action pair and the property of RKHS. Each sample of such kernelized state-action construct not only encapsulates the state-specific "local policy" (as in, which action was chosen in this state and what fitness value was observed correspondingly) but also generalizes across the neighboring state space and action space in a way that shares approximately the same policy structure: similar state configurations bonded with similar actions leads to similar fitness responses. Note that the similarity between any two actions and their derived quantities, (generalized) state-action pairs, can be established through the kernel function because, as mentioned in Section A, actions now take on their parametric, continuous representation (e.g., random vector) and thus are structurally like the state.

The policy-homogeneous state area, where a local policy generalizes as mentioned, is referred to as a *soft state* (Section IV.B) and is one of the key properties of the reinforcement field. If we further group similar decision contexts as such to form an action-oriented abstraction, the resulting clusters can help us identify a set of action-selection strategies, whereby, the agent can infer the best action at an unseen state by matching the samples of experience in a cluster comprising similar local policies with correlated fitness value responses. These clustered soft states can be thought of as a conceptual unit, referred to as *abstract action*, with which the agent derives the control policy (Section V.B).

Although the concept-driven policy is similar, in terms of its hierarchical structure, to the policy expressed in HRL methods, their policy representations are fundamentally different. Each abstract action in CDLA represents a set of coherent decision contexts that generalize a group of actions taken in different states (contexts) but nonetheless led to similar end results by virtue of a utility-like measure, fitness value. Therefore, the concept-driven action abstraction does not necessarily lead to sets of state sequences with spatially or temporally continuous properties as in other abstract action formulations such as the options framework in the context of Semi-MDP (or SMDP) [26] (which underlies most HRL approaches). Notably in these works, a so-called option, when applied in a particular state partition, would unfold in terms of a sequence of actions that lead to a terminal state within that partition boundary; the set of actions drives the agent step by step following a contiguous state area. Conversely, the concept-driven abstract action puts together a set of policy-embedded particles (or experience particles) that reference similar local policies as measured by their (correlated) fitness values.

By construction, each constituent "primitive" action (i.e., the action agent chooses to navigate the state space) in an abstract action (representing a decision concept) is not meant to be carried out in sequence like an option in a subgoal area. Instead, the action together with its paired state (within the same particle) is used to match other existing or "hypothetical" particles referencing other pairings of state and action. Note that these other particles can be either the existing ones in memory or purely hypothetical. Existing particles are those actually experienced by the agent as its decision processes unfold; hypothetical particles, by contrast, do not exist yet (likely unprecedented) but are formulated by combining hypothetical actions with the target states. Making such hypotheses is desirable as a skill because it allows for the RL agent to freely pair a state with any hypothetical actions for the purpose of policy inference. With the parametric-action model, the chosen action assumes a parametric form such that each hypothetical bond formed with a target state can be measured in terms of a degree of fitness, or compatibility, by a "critic" represented through *fitness value function*. The fitness value function is essentially an energy function [42]: the higher the fitness between the target state and a candidate action, the lower their energy, and hence, the higher the probability of choosing such action given its higher fitness degree with the target state.

The notion of abstract actions is similar in some sense to those in the SMDP homomorphism framework [27] that builds equivalent yet reduced MDP out of the structural symmetry and redundancy in the state space, which then leads to a form of policy symmetry. We also note that the policy homomorphism framework [12] explicitly represents such situation-specific control policies that generalize across the state space (or even across similar tasks).

Additionally, based on particle similarity and the formulation of hypothetical particles, we can further realize the operation of (decision-specific) *experience association* – a relatively simple instance of associative memory in psychology and cognitive science used to describe the ability to learn and memorize patterns, which are then later retrieved in response to relevant stimuli (e.g., for problem-solving) [37]. In fact, experience association is a key operation necessary to implement the update rule inherent in the particle reinforcement sampling algorithm as mentioned in Section A.

Experience association is useful potentially in complex domains with a large action space (involving multiple variables) compounded with non-stationary action dynamics, where applying the same action could lead to different consequences than are expected by the previously learned policy. For instance, in section IV, we will present a reinforcement field-based SARSA (aka RF-SARSA with the "A" being the parametric actions), in which the RL agent updates its working memory periodically such that the value prediction, as dictated by experience particles, also gets updated accordingly, reflecting the progressive (concept) shift of the underlying decision processes. In this adaptive SARSA policy learning, such memory association plays a crucial role in morphing the memory structure to reflect the latest system dynamics.

Furthermore, the concept-driven abstraction can be shown to reflect the correlation between decisions made across the state space and thereby, serve as a mechanism to reduce the decision points per state, which undoubtedly also relies on experience association. In Section V, we will detail a generalized SARSA variant, called G-SARSA, where the "A" now represents abstract actions. An abstract action, as a decision choice, is essentially a (functional) cluster of similar decision contexts grouped together as a conceptual unit. In this concept driven SARSA policy learning, the operation of memory association again plays a key role that renders clustering possible.

### C. The GRL Framework

The construction of reinforcement field and its further application, CDLA, are based on the following observations: i) actions are not merely atomic, immutable choices with behavioral details predefined but more so in practice, involve the local dynamics expressed as a continuous process ii) decisions made across different states often carry certain mutually-dependent relations and thus, correlated decisions can be associated with one another using their corresponding reinforcement signals (e.g., similarity and fitness) iii) correlated decision contexts can be further assembled together to represent a set of concept-driven abstract actions, each of which implicitly encodes a set of context-dependent local policies.

Based on these key aspects, this paper will conclude with an instance of the GRL framework, using the kernel-based approach, that periodically fulfill three interleaving processes: policy search, memory update, and functional clustering [31] to formulate concept-driven action abstraction. Subsequently, the agent uses abstract actions as decision choices to derive the control policy, further reducing the policy representation from the primitive action space to an often-smaller, arguably self-organizing, abstract action space.

On the fundamental level, the parametric action model enables the expression of action dynamics and its influence over the state space, thereby making it possible to establish the notion of decision contexts and their similarity. Through the property of the GPR and the RKHS, the kernel function plays a synergistic role in value prediction, experience association, and decision concept learning, without needing a complex set of inductive biases for seemingly different operations in the GRL framework. Working with the kernel machinery behind the scenes are the experience particles in the RL agent's working memory that can be characterized as a set of *functional data* [31] since each particle consists of a state-action pair coupled with its functional response, the fitness value (Section II.C). To discover coherent patterns hidden within such functional data set, GRL views the working memory as a graph such that particles can be clustered by a simple union of GPR [8] and the spectral clustering [16, 18] as they both share the same kernel matrix representation (more on Section V.A).

## II. PRELIMINARIES

We will first go through a few foundational steps before formalizing the notion of reinforcement field, which is the primary construct where the policy learning and generalization takes place in GRL. The policy learning in the field encompasses the following integrative processes: 1) resolving uncertainty in actions and their influence on the state space 2) collecting samples of experience as the decision processes move forward 3) updating the agent's working memory as guided by temporal differences 4) evaluating and improving the control policy via sample-driven value predictions, and 5) periodically tuning hyperparameters of the (learnable) similarity measure for incoming samples of experience, reflecting the most recent evidence in the working memory.

As a continuation of the last section, we have described these interleaving processes above in a relatively high-level form so that the focus can be on the GRL as a new policy learning framework, where the algorithmic components or parts of the system can be selectively changed according to the control domain of interest while the learning architecture and guiding principles remain the same. In this paper, we will develop a particular instantiation of the GRL framework using GPR as the backbone for value predictions, kernel functions as the basis for similarity learning, and state-action-reward-state-action or SARSA [4] as the baseline algorithm for learning a control policy under the formalism of Markov decision processes (MDPs). We will then generalize the baseline RL algorithm like SARSA with additional constructs including the parametric action model and the abstract-, decision concept-based policy learning that aim to better cope with fluid action behaviors and adapt to changes in the environment.

It is instructive to motivate the parametric action model that effectively transforms the regular, primitive actions in standard RL to a continuous, parametrizable representation that works with the kernel machinery within GPR – and at the same time can be easily captured and stored as a functional data set that constitutes the agent's working memory. Following each section also briefly outlines the thought processes leading up to the shift of perspective from the standard RL to the field-based policy learning approach.

### A. TD Learning and SARSA

Consider learning a policy over an MDP, where the RL agent perceives the current state $s \in S$, and chooses an action $a \in A$, followed by a transition into the next state $s'$. The agent subsequently receives a reward $r$ determined by a reward function $R: S \times A \to \mathbb{R}$. The state transition is governed by the environment dynamic represented through a map $T: S \times A \times S \to [0,1]$ that specifies the probability upon observing $s'$ by executing action $a$ in $s$.

Equivalently, $T$ can be interpreted as an Markovian transition operator that acts on a particular state-action pair $(s,a)$ and returns a probability distribution over the next state $s' \in S$ in terms of the set $\{P(s=s'|s,a) | s' \in S\}$. In particular, the conditional probability $P(s'|s,a)$ does not depend on the earlier history prior to the current state $s \in S$ due to the Markov assumption. The objective of the agent is to learn a control policy $\pi: S \times A \to [0,1]$ that leads to a maximized accumulated reward where $\pi$ effectively maps states to a probability distribution over actions. In particular, given a policy $\pi$, the corresponding value function $Q^\pi(s,a)$ computes the long-term accumulated reward in the form of the Bellman equation:

$$Q^\pi(s,a) = R(s,a) + \gamma \sum_{s' \in S} T(s,a,s') \sum_{a' \in A} \pi(s',a') \cdot Q^\pi(s',a') \quad (1)$$

However, very often the environment dynamic ($T$) and the reward function ($R$) are unknown to the agent, thus giving rise to the RL framework. Our description of policy learning in the reinforcement field will use SARSA [4] as the underlying control learner although other temporal-difference learning algorithms such as Q-Learning, TD-Lambda, etc., could be chosen instead. SARSA uses temporal differences to estimate Q-values (analogous yet different than the fitness value) with the following update rule, where $\alpha$ denotes a learning rate:

$$Q(s,a) \leftarrow Q(s,a) + \alpha \big[ R(s,a) + \gamma Q(s',a') - Q(s,a) \big] \quad (2)$$

Later in the introduction of reinforcement field, we shall see that Q-values estimated by (2) is used as guiding signals for estimating the fitness value of compound state-action pairs resulting from applying the parametric-action model in the state space. In CDLA, we further develop an even more compact policy representation, where the policy is derived instead through a set of abstract actions $A^c$ (as a higher-level representation of the primitive action set $A$ that are state-dependent). In this case, (2) will be generalized further as an update rule over abstract actions $A^{(i)} \in A^c$ (Section V).

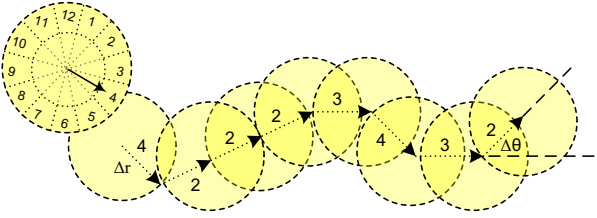

Fig. 1. Primitive actions parameterized by both a radius and an angle increment as a (constrained) random vector $(\Delta r, \Delta \theta)$ relative to the current position. Actions are indexed according to the clock direction. For instance, action 4, or $a^{(4)}$, leads the agent approximately in the 4 o'clock direction with both action variables subject to random variations.

The formulation of the action abstraction begins with defining action parameters, which are essentially a set of random variables that express the primitive actions of interest, $a \in A$, as a constrained random process.

*B. Parametric Actions*

To endow actions with richer expressions such that they are no longer just immutable decision choices with behavioral details predefined, we will introduce the notion of parametric actions as the first step to generalizing the standard RL framework. The rationale follows from the observation that, in complex control scenarios, executing an action involves a variational process over the configuration of the active state; that is, the state feature values change in response to the continuous process during which an action is applied.

The types of action-induced uncertainty could range from the unpredictability of an action when performed (e.g., errors from related actuators in robotic control) to the affordance or feasibility of an action with respect to a particular subset of states (e.g., task assignments to computational resources with service constraints), and certain localized effects an action may introduce to the environment such as consumptions or removals of objects. In particular, the localized effect induced by the action can be expressed in terms of the variation within a subset of state variables such that a group of states may share the same partial variations. For instance, STRIPS operators express such action effects through propositional logic and first-order predicate calculus. The action representation using DBN from the factored-MDP framework extends this concept further into stochastic actions. Specifically, an action in DBN triggers the state transition in terms of the probability conditioned only on a subset of state variables relevant to the predecessor state.

To further express the fluid aspect of the action with a dynamic descriptor as it acts upon a state, we shall see the parametric-action model offers certain advantages that go beyond the decision-theoretic perspective where actions are taken as logical choices. An example of applying the parametric-action model specifically to address localized effects can be found in the related work [6].

Parameterizing the actions introduces extra degrees of freedom necessary to express action dynamics. For clarity, an action choice ($a \in A$) from the standard RL is addressed as a primitive action. Again, we will refer to a primitive action in terms of the notation: $a \in A$ in contrast to an explicitly parameterized action $a(\mathbf{x}_a)$, where $\mathbf{x}_a$ is the action vector in the parametric action space $A^+$. In this manner, each primitive action is represented by a vector $\mathbf{x}_a$ with relevant variables that encode specifically what local variations can occur when it is chosen to act on a state. Fig. 1 illustrates a 2D navigation domain in which the actions of the agent are parameterized by an angle and a radius increment per move relative to the current position, where $\mathbf{x}_a = (\Delta r, \Delta \theta)$. For mnemonic purposes, we will also assume that the actions are indexed according to the clock direction: $\{a^{(i)} | i = 1 \sim 12\}$. For instance, taking the action $a^{(3)}$ moves the agent along the 3 o'clock direction with an appropriate step size while the underlying motion may be subject to uncertainty as the agent travels.

However, caveats arise when (primitive) actions possess these extra degrees of freedom because the parametric-action model effectively introduces an extended state space where the update rule in (2) no longer directly applies without a further extension. For ease of exposition, we will first define three special constructs that help in establishing the desired properties for the parametric-action model: i) augmented state space (Definition 1) ii) potential function over augmented state space (Definition 2) and iii) experience particles (Definition 3).

**Definition 1 (Augmented State Space)** *The augmented state space $S^+$ is a Cartesian product of the state space $S$ and the action parameter space $A^+$: $S^+ = S \times A^+$.*

Given Definition 1, we will denote an element of the augmented state space by $s^+ \in S \times A^+$, or simply, $s^+ \in S^+$, which is also equivalent to $(\mathbf{x}_s, \mathbf{x}_a) \in S^+$ obtained from peeking into the augmented state $s^+ = (\mathbf{x}_s, \mathbf{x}_a)$. Depending on the context, we use the vector notation $\mathbf{x}_s$ to denote a state $s \in S$ in the same way as the parametric form $\mathbf{x}_a$ associated with a primitive action $a \in A$. Similarly, we may also refer to $(\mathbf{x}_s, \mathbf{x}_a)$ as a "generalized" state-action pair to emphasize on the fact that both the state and the action are two entities that interact, or bond, with each other and results in an associated functional response (e.g., degree of fitness to be defined shortly). Broadly speaking, such entities can be embeddings of any complex structures (e.g., strings, pictures, graphs) or any objects for which there exist a valid kernel such that their inner product is well-defined. In this paper, we will assume that both $\mathbf{x}_s$ and $\mathbf{x}_a$ are in vector forms.

**Definition 2 (Fitness Function)** *A fitness function is a value function $Q^+: S^+ \to \mathbb{R}$ that maps an augmented state $s^+ \in S^+$ to a real number, referred to as the fitness value $Q^+(s^+)$ between the state vector $\mathbf{x}_s$ and the action vector $\mathbf{x}_a$ associated with the augmented state $s^+$.*

The fitness function in general can be thought of as a measure of compatibility, or bonding, between the state as a construct and the action as yet another construct, both of which can take on a complex form beyond scalars and vectors, although we will focus on the vector representation in this paper.

**Definition 3 (Experience Particle)** *An experience particle is an abstraction of a functional data point represented by the tuple $(s^+, Q^+(s^+))$, comprising an augmented state $s^+$ and its corresponding fitness value $Q^+(s^+)$.*

Note that $A^+$ in Definition 1 refers to the state space augmented to incorporate both the state as vectors $\{\mathbf{x}_s\}$ and the action as vectors $\{\mathbf{x}_a\}$, which is different from the action set $A$ comprising a set of primitive actions, now parameterized as $a(\mathbf{x}_a)$. The state vector and the action vector live in two separate vector spaces ($S$ and $A^+$ respectively), which in general, is not the same as simply combining two vectors to form a joint vector (although the joint vector can be a candidate representation for predictive purposes).

As we shall see shortly, it is also conducive as a parametric action representation to equip with a mapping from the parametric space $A^+$ to the primitive action set $A$, or more specifically, $f_A: A^+ \to A$. To accomplish this, we also need to allow for constraints specified on the action parameters such that each primitive action can be defined by action parameters, each assuming values that only fall within a certain range. One way to enforce such constraints is to limit the allowable value along each coordinate of the action vector according to a *yield function* [35] of the following form:

$$Y(x_{ai}) = P((x_{ai} + w_{ai}) \in \Gamma)$$

Under this notation, $x_{ai}$ represents a target value of the *i-th* coordinate of the action vector $\mathbf{x}_a$ while $w_{ai}$ represents a random variation along the same coordinate; $\Gamma$ is the set of acceptable values (or support). Further, one can control the probability threshold $\eta$ by which any random variations of $\mathbf{x}_a$ (due to $w_{ai}$) has a net result that always falls within $\Gamma$. This is compactly denoted as $(\mathbf{x}_a)_i = x_{ai} \mid Y(x_{ai} \geq \eta)$, where $\eta \in [0,1]$ is a threshold that determines the *η-yield region* that encompasses all acceptable "noisy values" of $x_{ai}$ for which the probability $Y(x_{ai})$ exceeds $\eta$. In practice, $x_{ai}$ corresponds to a "normal behavior" of the action along this coordinate. On the other hand, the uncertainty in how an action gets manifested is modeled by $w_{ai}$, which is chosen to follow a task-dependent probability distribution that is either estimated empirically or assumed to follow Gaussian distribution as a reasonable starting point. An action parameter value outside the boundary of $\Gamma$ is taken to be either infeasible or empirically improbable by assuming $\eta$ being close to 1.

In Fig. 1, the left-most circle indicates 12 possible orientations of movements (12 clock directions). Within each pie-shaped boundary between the inner and outer concentric circles represents a feasible range of the motion as determined by the action parameters $\mathbf{x}_a = (x_{a1}, x_{a2}) = (\Delta r, \Delta\theta)$. For instance, action 3 represents a motion moving along the direction to the east but with a tolerance of any errors within ±15 degrees. Similarly, the permissible radius is constrained to fall only within the inner and outer circles. In this example, $\mathbf{x}_a$ is chosen to model the expected local variation by which the related state variables (position) will shift upon a transition to the next state. If the current position of the agent is at $(u, v)$, then taking an action moves the agent further to $(u + \Delta r \cos \Delta\theta, v + \Delta r \sin \Delta\theta)$.

The 12 primitive action $\{a^{(i)} \mid i = 1 \sim 12\}$ can be compactly expressed by $a(\mathbf{x}_a)$, except that the action parameters in $\mathbf{x}_a$ is now also constrained to specific values associated with the 12 bounded areas as mentioned. Notationally, we will denote $\mathbf{x}_a^{(i)}$ to be the (indexed) action vector associated with the primitive action $a^{(i)}$. Under this action model, however, a careful distinction must be made between action parameters and action vectors, which, for notational convenience, are both denoted by $\mathbf{x}_a$. Action parameters are random variables with values yet to be determined according to their underlying probability distribution(s) and are subjective to the feasible range-specific constraints. Since a primitive action is in general represented by multiple parameters, we reserve the term, action vector, for a particular realization of action parameters with their ordering fixed and consistent, and therefore can be treated as random vectors.

The precise expression for the constraints can either be obtained empirically or specified by the domain expert as a modeling assumption. For instance, assuming there is a perfect symmetry in how the navigational agent behaves in all directions, one can set $x_{a1}$ (or $\Delta r$) to be always constrained between the inner and outer circles, and $x_{a2}$ (or $\Delta\theta$) to take on bounds $\{[\frac{\pi}{12} \cdot (i-1), \frac{\pi}{12} \cdot (i+1)) \mid i = 1, \dots 12\}$, i.e., 12 angles each spanning $\pi/6$ with the index $i$ conforming to the clock direction. For instance, the 3-o'clock action $a^{(3)}$ has a feasible range within $[\frac{5\pi}{12}, \frac{7\pi}{12})$ with an average orientation at $\frac{\pi}{2}$, pointing exactly at the 3 o'clock direction.

By introducing action parameters, complex actions (especially those involving multiple variables) can be relatively easily modeled and thereby incorporated into the policy search. In particular, the action behaviors can be allowed to vary over time reflecting true scenarios in practice. For instance, a robotic arm may not always follow the expected trajectory induced by the forward kinematics to reach a particular spatial region due to small errors in the prescribed axis- or angle-related parameters specifying joint links and their connections. Similarly, a robot navigates in an open area with uneven terrain may not always take a step with a fixed length or resolve its orientations without unexpected tilts. Additionally, the policy control on the level of primitive actions (as in standard RL) technically does not have a solution when action availability is not a "constant" (e.g., imagine the agent get stuck and suddenly a subset of navigational choices become infeasible) because a change in action set effectively redefines the underlying RL problem.

With the parametric action model, the fitness value at different configurations of the augmented state space can always be estimated (under the condition of sufficient exploration), and by mapping the parametric action back to the primitive action set, the agent can then select an alternative action based on its value estimate. Periodic updates of the (fitness) value function will then gradually favor towards the new action based on the progressively updated working memory in which this new alternative consistently dominates the others (Section IV). Overall, the control policy needs the capacity as such to adapt to unexpected, fluid action behaviors and thereby exhibit some level of fault tolerance to be practically applied in complex, real-world applications.

### C. Representing Evidence as Functional Data

Under the GRL framework, the agent's working memory comprises a set of samples of experience coupled with a learnable similarity measure that (implicitly) draws a connection from one sample to another, structurally forming a graph while operationally set for the memory association that serves as both the sample update rule, as new experience comes in, and as a mechanism for policy inference. To endow the memory with these properties, we need a flexible, self-organizing structure to represent the evidence.

With the parametric action model, every time when the agent executes an action in a state, an experience particle (Definition 3) is created, encapsulating the corresponding augmented state along with a "measure" of its fitness, referred to as the fitness value (Definition 2). To illustrate, suppose that $\mathbf{x}_s$ and $\mathbf{x}_a$ respectively represent a state vector and an action vector, resulted from executing a primitive action $a \in A$. The corresponding augmented state $s^+$ is an element of $S \times A^+$ given by $(\mathbf{x}_s, \mathbf{x}_a)$. For notional convenience, we will assume that $\mathbf{x}_s$ has dimension $m$ and $\mathbf{x}_a$ has dimension $n$, and write $\mathbf{x} = (x_{s1}, \ldots x_{sm}, x_{a1}, \ldots x_{an})$ as a joint vector of $(\mathbf{x}_s, \mathbf{x}_a)$ (while bearing in mind that state and action are two separate entities, each with its own vector space). The fitness function $Q^+$ evaluated at this augmented state therefore can be expressed as:

$$Q^+(s^+) = Q^+(\mathbf{x}) = Q^+((\mathbf{x}_s, \mathbf{x}_a)) \qquad (3)$$

We will also denote $q_i$ as the output of evaluating $Q^+(\cdot)$ at the $i$-th training instance $\mathbf{x}_i$. Given (3), if the agent traverses the state space by $k$ steps, a sequence of particles of size $k$ will have been generated and retained in the memory:

$$\Omega: \{(\mathbf{x}_1, q_1), (\mathbf{x}_2, q_2), \ldots, (\mathbf{x}_k, q_k)\} \subseteq X \times \mathbb{R},$$

where $X$ represents the set of augmented states in the joint vector-form.

It is helpful for the moment to consider the working memory (denoted by $\Omega$) as a *functional data set*, where each experience particle is also associated with a fitness value. How these fitness values are estimated is deferred to the next section. Under the functional data perspective, each experience particle, holding a snapshot of both state and action with stochastic effects resolved, can be interpreted as a local policy that generalizes into the neighboring (augmented) state space through properties of RKHS to be discussed shortly in Section III. In brief, this local policy generalization is made possible by allowing the actions to take on a parametric form underneath while retaining their original identities as decision choices. Having an action-induced state representation, i.e., the augmented state space, along with kernel function as correlation hypothesis, we now have a basis for realizing associative memory in which augmented states can be related to one another through their inner products in the kernel-induced feature space [1], [15]. Representing the policy learning experience as functional data also facilitates a more compact representation of the underlying state space that otherwise would become much larger due to complex action dynamics.

One implication for $\Omega$ being a functional data set is that the similarity between any two data points now needs to be defined by considering both the data (e.g., augmented states) and their functional responses (e.g., fitness values). Since the RL agent's memory $\Omega$ essentially represents a learning experience for making better decisions, we will formalize the notion of "decision similarity" factoring in state, action – and their fitness after we introduce GPR in the next section. The functional similarity serves as an instrument for implementing an associative memory that relates similar decisions made in the past as they are gauged by their fitness values. Facing a new decision point where the best action is to be determined, the agent has the capacity to make hypotheses (hypothetical state space) by replacing the "action slot" of the existing particles with alternative actions and inferring the fitness values for these hypotheses (Section IV.A).

To demonstrate this decision-specific concept learning using the property of functional data, we will examine a task assignment domain in the section of empirical study. As a brief illustration, consider the case where a continuous stream of user tasks is to be assigned to a resource pool (e.g., a cloud computing network [21]) comprising several servers, each of which has a time-varying resource profile. Computational resources are shared by a group of users and hence their loadings change over time. The primary objective for an ideal task-assignment policy is to identify the appropriate servers matching user tasks in real-time so as to optimize a given performance metric (e.g., minimum turnaround time). In this scenario, choosing the actions representing possible assignments to servers using a standard RL formulation would likely fail to reach a stable policy due to the time-varying properties of the target servers. As mentioned earlier, the commonly used multiagent approach based on the standard RL formulation often inevitably introduces combinatorial state and action spaces jointly having a large feature space, making it difficult to estimate utility values under higher population of agents (e.g., servers).

On the other hand, if we shift our perspective by modeling each task-assignment choice through a set of action parameters that are dynamically changing but shared across all servers (each with appropriate constraints), then through functional similarity between any task-assignment decisions as measured by the kernel function, we will be able to learn (approximately) optimal task-assignment policy regardless the server population (i.e., the size of the resource pool). For instance, the percentage CPU time, available memory, etc., are among the candidate action parameters for it is these attributes that determine the server usability and efficiency. In this manner, the primitive action set effectively models all the currently available servers through the shared action parameters, leaving the state space to capture incoming tasks (e.g., job types, CPU demand, etc.) in addition to perhaps some global statistics necessary to derive a human-readable task-assignment policy.

Moreover, the set of available servers may vary with time in addition to the time-varying resource profile. In such case, the set of action parameters remain invariant and thus the policy derived with respects to these shared action parameters (modeling the action of assigning tasks to various servers) will continue to apply regardless of the shift in the action set (i.e., varying server availability) or action semantics (i.e., varying resource profile).

The key in combining the state space with a shared action parameters is to identify the functional relationship that maps state variables and action parameters, to their corresponding fitness values, which then implicitly defines local policies for the experience particles involved. It will later be substantiated in Section IV that the state-action pairs $\mathbf{x}^{(i)} = (\mathbf{x}_s^{(i)}, \mathbf{x}_a^{(i)})$ in $\Omega$ when embedded within kernel functions can effectively "loosen" the boundary between a state and an action, leading to a more compact yet fluid form of state transitions, referred to as "soft" state transitions.

Equipped with the above sample-based functional constructs, we need one more step to establish the reinforcement field, which is to represent the value function in a manner that can incorporate both state and action vectors (as two separate mathematical objects) – and in the meantime, output a value estimate for the particle to take on such state-action combination (bonded as a single object). The value estimate as such is interpreted as fitness for the bonding between the state and the action, as both entities, in principle, can be arbitrarily large structures with vector representation being the special case. The fitness function is similar to the role of value functions in standard RL but is envisioned to take into account more general interactions between states and actions. This version of GRL represents the fitness value function by a progressively updated Gaussian process.

### D. Gaussian Process Regression

Consider the training data $\Omega: \{(\mathbf{x}_1, q_1), (\mathbf{x}_2, q_2), \ldots, (\mathbf{x}_k, q_k)\}$ given earlier. The goal is to identify a function $Q^+(\cdot)$ fitting these particles with a tradeoff between quality of the prediction (i.e., data fit) and smoothness assumptions of the function (i.e., model complexity) [8]. Value predictions using GPR first assume a normal prior distribution over functions, and subsequently reject those not consistent with the observations, which in our case are the experience particles. Suppose that we wish to reconstruct the target function $Q^+$ from the noisy observations of fitness values, $q_i = Q^+(\mathbf{x}_i) + \varepsilon_i$, where $\varepsilon_i$ assumed to be an i.i.d. Gaussian noise such any subset $\{q_i\}$ follows a Gaussian distribution with a kernel or covariance function $k(\cdot,\cdot)$. The prior distribution over the observed values $\{q_i\}$ is thus given by: $\mathbf{q} \mid \mathbf{x} \sim N(0, K)$, where $K_{ij} = k(x_i, x_j)$ and the fitness values are assumed, for simplicity, to be centered around a zero mean.

In this paper, we will explore two different kernels – an SE kernel in (4) and a product kernel in (5). Equation (4) is a noisy squared exponential kernel (SE) that combines all the state and action variables in the exponential term, i.e.

$$k(\mathbf{x}, \mathbf{x}') = \theta_0 \exp\left(-\frac{1}{2}(\mathbf{x}-\mathbf{x}')^T D^{-1}(\mathbf{x}-\mathbf{x}')\right) + \theta_1 \sigma_{noise}^2 \quad (4)$$

In (4), the diagonal elements $D_{ii}$, signal amplitude $\theta_0$ and the noise magnitude $\theta_1$ correspond to the hyperparamters of the kernel. Alternatively, the state vector $\mathbf{x}_s$ and the action vector $\mathbf{x}_a$ can be represented via two separate kernels to distinguish them as two different objects; this gives:

$$k(\mathbf{x}, \mathbf{x}') = \theta_0 k_s(\mathbf{x}_s, \mathbf{x}_s') k_a(\mathbf{x}_a, \mathbf{x}_a') + \theta_1 \sigma_{noise}^2 \quad (5)$$

where $k_s(\cdot,\cdot)$ and $k_a(\cdot,\cdot)$ represent the state kernel and action kernel, respectively. After observing a sequence of experience particles, the predictive distribution for the value of a novel augmented state $\mathbf{x}_*$ is given by: $q_* \mid X, \mathbf{q}, \mathbf{x}_* \sim N(\bar{q}_*, \sigma)$ where,

$$\bar{Q}^+(\mathbf{x}_*) = \bar{q}_* = \mathbf{k}(\mathbf{x}_*, X)^T K(X, X)^{-1} \mathbf{q} \quad (6)$$

$$\sigma = k(\mathbf{x}_*, \mathbf{x}_*) - \mathbf{k}(\mathbf{x}_*, X)^T K^{-1} \mathbf{k}(X, \mathbf{x}_*) \quad (7)$$

In (6) and (7), the set of augmented state vectors as training examples are stacked in $X$ as column vectors for convenience; thus, $\mathbf{k}(\mathbf{x}_*, X)$ represents a column vector with each entry evaluated by $k(\mathbf{x}_*, \mathbf{x}_i)$. In particular, the mean prediction following (6) is used for evaluating the fitness value $Q^+$ for a given augmented state.

The choice of using GPR as a representation for the fitness function boils down to the consideration of a few desired properties. First, the GP-based fitness function assimilates both the state features and action parameters through the kernel as a covariance function such that the similarity in value predictions also reflects the degree of similarity of state-action combinations. Secondly, since the value prediction is driven by the samples (i.e., experience particles) having certain correlation to one another (measured by the kernel), the policy can theoretically be deduced everywhere in the state space including those regions not yet being explored. That is, if there exist sufficiently correlated particles in the proximity, the kernel as a correlation hypothesis for particles will continue to apply without requiring the agent to explicitly explore the unknown region. We shall see that this property along with the action operator (Section III.A) have benefits that build upon each other.

Further, the kernel in GPR adapts to the ongoing variations in the environment dynamics by the ARD procedure [8] that (periodically) tunes the kernel hyperparameters using conjugate gradient optimization. In particular, the ARD procedure performs an automatic model selection by maximizing the marginal likelihood of the data:

$$\log p(\mathbf{q} \mid X) = -\frac{1}{2}\mathbf{q}^T K^{-1}\mathbf{q} - \frac{1}{2}\log|K| - \frac{n}{2}\log 2\pi, \quad (8)$$

where entries in $K$ are evaluated through a given kernel (e.g.,

(4), (5)). The term involving the observed fitness values $-\frac{1}{2}\mathbf{q}^T K^{-1}\mathbf{q}$ represents the data fit given the evidence (particles) collected from the policy learning, while the term $-\frac{1}{2}\log|K|$ represents the complexity of the fitness-value model $Q^+(\cdot)$. The ARD process corresponds to maximizing (8) by, for instance, incorporating a separate hyperparameter for each variable of the augmented state (i.e., state variables and action parameters). This helps in determining the relevance degree that each variable contributes to the final prediction. To determine the optimal hyperparameters, first express explicitly the marginal likelihood function (8) conditioned on the hyperparameters as $\log p(\mathbf{q}|X,\theta)$, followed by taking partial derivatives of (8) with respect to all hyperparameters to give:

$$\frac{\partial}{\partial \theta_i}\log p(\mathbf{q}|X,\theta) = \frac{1}{2}\mathbf{q}^T K^{-1}\frac{\partial K}{\partial \theta_i}K^{-1}\mathbf{q} - \frac{1}{2}tr\left(K^{-1}\frac{\partial K}{\partial \theta_i}\right) \quad (9)$$

By (9), the complexity of using ARD to determine the relevance degree for each augmented state variable toward the value prediction $Q^+$ is largely dominated by the matrix inversion of $K$. We shall see in Section IV an instance of particle reinforcement sampling algorithm using the property of temporal differences to preserve only the essential experience particles necessary for the policy inference, thereby reducing the dimension of covariance matrix for computational advantage.

### E. Control Policy with Kernelized Augmented State

Since the objective of the fitness function $Q^+$ is to act as a critic for experience particles, the policy is expected to take on the form of a functional depending on $Q^+$ i.e., $\pi[Q^+]$. Using a *softmax function*, the policy over a pair of state and a parametric action (referenced by a particle) can be represented by:

$$\pi(s,a^{(i)}) = \frac{\exp\left[Q^+\left(s,a^{(i)}\right)/\tau\right]}{\sum_j \exp\left[Q^+(s,a^{(j)})/\tau\right]}$$
$$= \frac{\exp\left[Q^+\left((\mathbf{x}_s,\mathbf{x}_a^{(i)})\right)/\tau\right]}{\sum_j \exp\left[Q^+\left((\mathbf{x}_s,\mathbf{x}_a^{(j)})\right)/\tau\right]} \quad (10)$$

In (10), we have used $(\mathbf{x}_s,\mathbf{x}_a^{(i)})$ to denote the augmented state associated with the state-action combination $(s,a^{(i)})$ at the level of primitive actions $a^{(i)} \in A$. On the other hand, $\tau$ denotes a temperature that models the confidence for the control policy: a low temperature results in a "harder" distribution over actions, with which the agent has a relatively high confidence in choosing the best action, whereas a high temperature leads to a "softer" distribution, with which the "best" action is relatively ambiguous (since the probability mass is distributed more evenly across different actions).

Since the augmented state $(\mathbf{x}_s,\mathbf{x}_a^{(i)})$ in (10) is defined to be an element of the Cartesian product $S \times A^+$, an appropriate kernel-dependent transformation is necessary before such representation can be applied in the context of GPR. For instance, to use the (noisy) SE kernel in (4), the augmented state can be first transformed into a joint vector $\mathbf{x} = (x_{s1},...x_{sm},x_{a1},...x_{an})$. In this manner, the similarity between any two augmented states can be evaluated through the quadratic form within (4); that is, $(\mathbf{x}-\mathbf{x}')^T D^{-1}(\mathbf{x}-\mathbf{x}')$, which amounts to a weighted combination between the state and action vector components (with weights being the kernel's hyperparameters). Consequently, for notational convenience, $Q^+((\mathbf{x}_s,\mathbf{x}_a))$ is simplified to $Q^+(\mathbf{x})$, where $\mathbf{x}$ in general represents a combined object between the state and the action in a general sense such as a joint vector as mentioned.

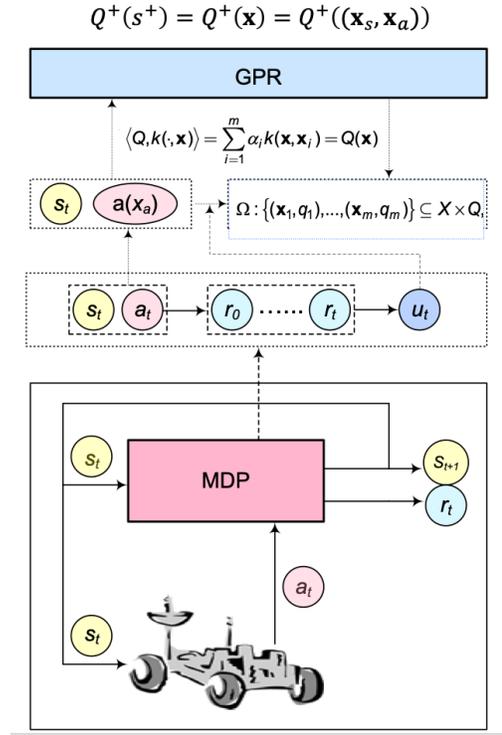

Fig. 2. Fitness value prediction using a hierarchical learning architecture where the agent is depicted as a rover navigating on a surface. The base layer learns the control policy under MDP where primitive actions apply. Primitive actions are then "expanded" into their parametric forms and later combined with the states along with the base-level value estimates to form experience particles, which are periodically fed into the GPR layer while being updated in parallel to incorporate new learning experience.

The representation for the *kernelized augmented state*, $k(\mathbf{x},\cdot)$, depends on the chosen kernel function $k(\cdot,\cdot)$, and thus in principle can assume any functional form (e.g., graph kernel with $\mathbf{x}$ encoding a graph) besides a simple joint vector within an SE kernel. Therefore, so long as the chosen kernel is valid (i.e., a Mercer kernel [1] that always corresponds to a valid inner product), then the similarity between any two kernelized augmented states can be evaluated through the inner product $k(\mathbf{x},\mathbf{x}')$ in the kernel-induced feature space.

To estimate the value of $Q^+$ for each augmented state, we can adopt a simple two-tier architecture where the GPR works in parallel with an underlying control learner such as SARSA that "perceives" parametric actions $a(\mathbf{x}_a)$ as individual choices (i.e., primitive actions $a \in A$) without knowing the underlying parameters $\mathbf{x}_a$ that govern the action dynamics. Recall from Section II.B that a mapping from the primitive action set $A$ to the parametric action space $A^+$ can be established by specifying appropriate constraints for action parameters. In other words, we wish for the GRL agent's control learner (e.g., using the SARSA update rule) to inherit the same iterative value iteration approach using primitive actions as in the standard RL formulation. In the case of SARSA, this effectively corresponds to the update rule in (2). Other TD learning methods such as Q-learning is equally applicable as well.

To retain the MDP formalism at the level of the SARSA engine (i.e., the control learner interpreting actions in the sense of decision choices, or primitive actions) while enabling the fitness value estimation at the level of parametric action model, we consider a hierarchical approach using GPR as illustrated in Fig. 2. First, we obtain a utility estimate for a given state-action pair $(s, a)$ using SARSA in the base layer. Subsequently, this utility estimate propagates "upward" to the GPR layer to serve as a training signal for the associated augmented state, which is the compound state-action representation that the base layer does not need to be aware of. At the level of GPR, we maintain a working memory $\Omega: \{(\mathbf{x}_1, q_1), \dots, (\mathbf{x}_k, q_k)\}$, where $\{q_i\}$ is essentially the utility estimates progressively obtained from the base layer using value iteration. How does the information learned from the base-level decision processes generalize to the larger augmented state space? This is where we need properties of the GP and its associated kernel function (Section III).

As mentioned in Section II.B, an augmented state $s^+ \in S^+$ is obtained from resolving the stochastic effects when an action is applied at the state space $S$ (also referred to as action dynamics). The stochastic effects are captured by the action parameter space, through which the target (primitive) action essentially gets transformed into a constrained random vector. Sampling from this random vector results in a realization of the action in vector form $\mathbf{x}_a$ (now a constant vector) that serves as a constituent part of the augmented state. The aforementioned process, i.e., the resolution of action dynamics, can be represented by a task-dependent primitive-to-parametric map $f_{A^+}: A \to A^+$ that goes from the primitive action $a^{(i)} \in A$ to its corresponding action parameters $\mathbf{x}_a^{(i)} \in A^+$, in which the index $i$ implies appropriate constraints being applied to the underlying parameters consistent with the feasible range for the indexed primitive action $a^{(i)}$ (see an example in Section II.B).

One side benefit for this hierarchical approach is that the complexity for fitness value estimation does not need to be constrained by the size of the state space. For instance, the SARSA engine can be chosen to operate on a set of discretized state partitions $S^{(k)} \subseteq S$ so that each fitness value for an augmented state is effectively also a utility estimate, albeit with lesser precision, when an action is applied in the underlying state partition. Each state partition $S^{(k)}$ maintains a reference to several particles, each of which has an assigned utility estimate, reflecting the subtle differences in the exact values of the augmented states. Specifically, this amounts to the partition $S^{(k)}$ referencing the set $\{(\mathbf{x}_s, \mathbf{x}_a)_j^{(k)}\}$ with $(\mathbf{x}_s)_j^{(k)}$ falling into the boundary of $S^{(k)}$ where the subscript index distinguishes different particles. Doing so implies that there is a trade-off between the accuracy in the utility estimate, the granularity of the underlying state partition, and the particle density: the finer the partition, the higher the particle density, and hence the higher the precision of the value estimate.

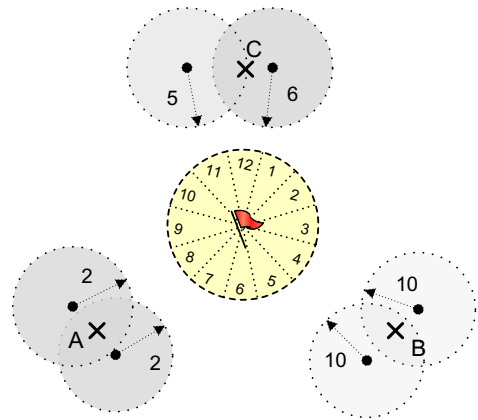

Fig. 3. Illustration of policy inference in the proximity of states where the local control policies are known a priori. Query points such as A, B and C are marked by crosses. Point A is within the scope of the decision contexts (states) where the 2-o'clock action is optimal while point B is within the nearby states where the 10-o'clock action is optimal. Point C, however, is being influenced stronger by the state to the right, which would drive the agent at C to favor the 6-o'clock direction over 5 o'clock.

## III. REINFORCEMENT FIELD

### A. Action Operator

The reinforcement field combines both the insights from various field theories in physics as well as properties of the kernel function. In the most abstract sense, a field acts on an object (e.g., a particle) in a manner governed by the law of the field. For instance, a charged particle moving in a magnetic field is subject to a Lorentz force that induces a circular motion on the particle. The magnetic field implicitly "prescribes" a particular state sequence for the particle to follow during the course of its travel. The circular motion can be justified by the principle of least action [7] in which an *action*, for the case of one particle, is defined in terms of the Lagrangian $L$ of the form: $\int_{t_1}^{t_2} L(q(t), q'(t)) dt$, where $q$ refers to a generalized coordinate of a physical quantity of interest such as the position of the particle. The action in this formulation essentially represents a global constraint for the particle trajectory. Incidentally, the term "action" overlaps with that in the RL terminology. Minimizing the action with respect to $q(t)$ (and $q'(t)$) using calculus of variation leads to the Euler-Lagrange equation that effectively represents the local dynamics of the particle, i.e., the motion equations for the particle. Thus, we see that in a physical system, there exists a dynamic process that drives state transitions systematically towards a certain objective (e.g., shortest path of travel) in a manner consistent with the law of the field.

By connecting this analogy to the GRL framework, the objective of maximizing the accumulated reward corresponds to a global constraint while the local dynamics can be expressed by the parametric actions. An action in this sense assumes the property of an operator, often in a functional form (e.g., the integral operator with the Lagrangian $\int L(\cdot)dt$) over a set of related parameters (e.g., position and velocity). For the reasons stated above, we seek to generalize the parametric action further into an operator-like construct with tunable parameters characterizing the local dynamics of state transitions and simultaneously links it to the fitness function.

As an example, consider three snapshots in a 2D navigation domain in Fig. 3 where the motion of the agent is again parameterized by the radius and angle increment i.e., $(\Delta r, \Delta \theta)$. Both $\Delta r$ and $\Delta \theta$ are continuous random variables, each with its feasible values constrained by corresponding yield functions as discussed earlier in Section II.B. Given this simple parametric action model, one can alternatively represent the state transition in terms of a matrix multiplication:

$$\begin{bmatrix} 1+\frac{\Delta x}{x} & 0 \\ 0 & 1+\frac{\Delta y}{y} \end{bmatrix} \begin{bmatrix} x \\ y \end{bmatrix} = \begin{bmatrix} x+\Delta x \\ y+\Delta y \end{bmatrix}, \tag{11}$$

where $\Delta x = \Delta r \cos(\Delta \theta)$ and $\Delta y = \Delta r \sin(\Delta \theta)$. In this case, action is represented as a matrix, a linear operator, that takes an input state $(x, y)$ and produces the next state $(x+\Delta x, y+\Delta y)$.

Recall from Section II.B that we had used the notation $a(\mathbf{x}_a)$ to denote the parametric action, which, depending on the constraint imposed on the parameter $\mathbf{x}_a$ (as a random vector), would correspond to a particular primitive action. We can endow $a(\cdot)$ the property of an operator because, if we can fix the argument $\mathbf{x}_a$ (e.g., through a sampling process such that $\Delta x$ and $\Delta y$ are determined), then $a(\mathbf{x}_a)$, in this case, is essentially a function in the 2-by-2 matrix form that produces a mapping from an arbitrary start position to its successor position, reflecting a single step of an action with movements specified according to the sampled values of $(\Delta x, \Delta y)$.

There are potentially several ways to formalize the notion of action operator, which can be a linear operator such as matrices but in general should allow any arbitrary mappings including non-linear operations. We will first consider an action operator as a higher-order generative function $f_a: \mathbf{x}_a \to a(\mathbf{x}_a)$ that maps the action parameter $\mathbf{x}_a \in A^+$ to another function $a(\mathbf{x}_a): s \to s'$ that acts on an input state $s$ and produces an output state $s'$ according to the value of $\mathbf{x}_a$, where $\mathbf{x}_a$ serves as the constraint as for what successor states $s'$ can be. Note that we have overloaded the notation $a$, originally reserved for the primitive action, to take on a new meaning of a function, $a(\cdot)$, that operates on states. $f_a$ is a higher-order function because its codomain again consists of functions; it is also generative in the sense that, depending on the action parameters and their scopes, $f_a$ produces different actions, as functions, each of which acts on the state differently (e.g., the 12 parametric actions representing clock directions as we have seen in previous sections).

Broadly speaking, we can endow $f_a$ with a more general mapping, specifically through $a$ (or $a(\cdot)$), such that, under appropriate parameterization $\mathbf{x}_a \in A^+$, $a(\mathbf{x}_a)$ maps from $\mathbb{R}^m$ to $\mathbb{R}^n$; i.e., $a(\mathbf{x}_a): \mathbb{R}^m \to \mathbb{R}^n$, for which the output dimension $n$ can be different from the input dimension $m$. This is a useful abstraction, for instance, when we consider taking an action in a transformed state space in $\mathbb{R}^n$, as an approximation to the original state space in $\mathbb{R}^m$ (where $m > n$), so that the credit assignment problem in the smaller state space is computationally less expensive to solve. Effectively, the two-tier learning architecture proposed earlier in Section II.D for the 2D navigation domain is an application of such action operator if we consider the action parameters $(\Delta r, \Delta \theta)$ as extra state variables being abstracted away from the "base" state in 2D, i.e., $(x, y)$, and hence are hidden from the MDP layer (i.e., the base layer) – while the GPR layer perceives the full, augmented state, which can be "represented" as a joint vector in 4D $(x, y, \Delta r, \Delta \theta)$ that gets fed into the kernel function. Note that such a 4D representation is more of an engineering choice than for mathematical rigorousness because the augmented state space is really the Cartesian product between two independent vector spaces (Definition 1), each in 2D:

$$(\mathbf{x}_a, \mathbf{x}_s) = ((x,y), (\Delta r, \Delta \theta)) \in S \times A^+.$$

It is noteworthy to consider a special case where the mapping $a(\cdot): \mathbb{R}^m \to \mathbb{R}^n$ holds with $n = 1$. This occurs when the action is considered as a function that maps a state in $\mathbb{R}^m$ to a real number such as a reward. One use case for such action operator is the contextual bandit problem [38], which is a special case of RL (and sometimes referred to as associative or one-step RL [39]). Contextual bandit is often used to model recommender systems in which items are selected according to the preference of the users. The action corresponds to the selection of items and the state (context) is modeled by the user attributes such as purchase history, geolocation, and demographic information, among many other attributes that are helpful in establishing the user profile. Previously in Section II.C, we previewed a task-assignment problem to be illustrated in the empirical study (Section VI), which has a similar structure as the recommender system since both tasks share the same high-level match-making operation – i.e., finding the best object in category X to match another object in category Y – and therefore, can be represented by the same family of action operators:

$$a(\cdot): \mathbb{R}^m \to \mathbb{R}.$$

However, the task-assignment problem has an extra complication: the resource profile (e.g., CPU load, memory), due to the condition of user sharing, varies over time, potentially rapidly, unlike most attributes typically considered for user profile in recommender systems, which are relatively static with respect to the recommendation requests (e.g., demographic information usually does not change as the user browses the inventory and makes purchases online through the recommender system). Therefore, a dynamical system like task assignments (or resource allocations) would fit into the parametric action model more closely especially when we consider the whole GRL machinery with action dynamics being captured by parametric actions coupled with the periodic update mechanism inherent in the working memory and fitness value predictions.

The notion of action operator helps in conceptualizing reinforcement field in the sense that how the "field" acts on the (experience) particle can be captured by the notion of operator continuously shifting the state configuration, which equates to the agent following in a specific path in the state space to fulfill an optimization objective (to be discussed shortly). This is in many ways analogous to, say, the electromagnetic field acting on a charged particle such that its internal state changes over time according to the path of least action as an optimization objective, where the shifts of the particle state can be explained and predicted by the Lagrangian operator [40].

Note that the action operator $f_a$ effectively generalizes the parametric-action model to broader usage scenarios where different mappings $a(\cdot)$ may be required to cope with different categories of control tasks. With the given action parameters fixed (that effectively serve as constraints), we may wish to consider the action as: i) a class of functions that map from one state to another within the same state space; e.g., the 12 clockwise actions in Fig. 1 ii) a class of functions that map from one state space to another space potentially of different dimensionality (and therefore, of different state space abstraction, say, for computational reasons), and iii) functions that formalize contextual bandit problems by mapping a state (or context) to a reward.

With the action operator, the objective of the learning in standard RL is altered slightly to the following statement: *determine the action operator that acts on each state in a manner that leads to a trajectory for a maximum accumulated reward.* Effectively, the action operator serves as a local constraint, through the action parameters, that each state transition must follow as it alters the state configuration. The fitness function, on the other hand, measures the value of the action operator as it acts on a state, again, through the action parameters. Overall, the relationship between parametric actions and primitive actions remains the same as before; however, the operator perspective provides us a functional and dynamical view as for how taking an action gets carried out in the state space, and perhaps draws a better connection to how a reinforcement field is constructed to be discussed next.

### B. Policy Learning in the Field

Recall from (6) that the fitness value function is represented through a linear combination of the kernels centered at different augmented states weighted by the observed fitness values. Without explicitly specifying the test augmented state, (6) can be rewritten in the following form:

$$\bar{Q}^+(\cdot) = \sum_{i=1}^{m} \alpha_i k(\mathbf{x}_i, \cdot) \qquad (12)$$

where $\boldsymbol{\alpha} = K^{-1}\mathbf{q}$. Since each $k(\mathbf{x}_i, \cdot)$ in (12) effectively maps an input pattern $\mathbf{x}_i$ (an augmented state) into a function, therefore taking a linear combination of these kernels forms a vector space. Consequently, $\bar{Q}^+(\cdot)$ in (12) can be treated as a vector in a vector field (of functions) with $k$ as the *representer of evaluation*. Specifically, $\bar{Q}^+(\cdot)$ is a vector in Reproducing Kernel Hilbert Space (RKHS) [15] with the reproducing property: $Q^+(\mathbf{x}) = \langle Q^+(\cdot), k(\cdot, \mathbf{x}) \rangle$ where $\langle \cdot, \cdot \rangle$ denotes an inner product evaluated in the unique RKHS associated with the kernel $k(\cdot, \cdot)$ (also see Moore-Aronszajn theorem [28] for more details). By virtue of this property, the known experience particles effectively establish a field through which a novel particle obtains its fitness value by comparing against all the other particles. Each comparison evaluates a particular degree of correlation to a known particle. The closer the test point (i.e., the new augmented state in query) is to a known particle (or more specifically, the augmented state referenced by the particle), the closer their fitness values are. Consequently, the final prediction is governed by those particles in the proximity of the test point. In addition, by Mercer's Theorem [1], any positive-definite kernel assumes an expansion of its eigenfunctions $\boldsymbol{\phi}$; that is,

$$k(\mathbf{x}, \mathbf{x}_*) = \sum_{i=1}^{N} \lambda_i \phi_i(\mathbf{x}) \phi_i^*(\mathbf{x}_*) \qquad (13)$$

where $N$ approaches to infinity if the kernel is non-degenerative such as the SE kernel in (4). Note that, for the augmented state vector $\mathbf{x}_* \in \mathbb{R}^n$, $\phi_i^*(\mathbf{x}_*)$ in (13) is identical to $\phi_i(\mathbf{x}_*)$. Equation (6) together with (13) entails that using GPR with a non-degenerative kernel is equivalent to a Bayesian linear regression model using infinitely many basis functions. This effectively implies that the fitness value prediction for a new particle will be bounded by a relatively small variance (see (7)) so long as certain experience particles exist in the neighborhood such that their associated augmented states are sufficiently correlated. We are now ready for the definition of a reinforcement field.

**Definition 4 (Reinforcement Field)** *A reinforcement field is a vector field in Hilbert space established by one or more kernels through their linear combination as a representation for the fitness function, where each of the kernel centers around a particular augmented state vector.*

As suggested by (11) an action operator is a function of both the state features and action parameters. As a result, the exact form of the action operator is not determined until all the parameters in the operator are fixed. Once the state of interest is determined along with the choice of the (primitive) action $a \in A$, the action operator then acts on the state such that it resolves the stochastic effects modeled by the action parameters into a fixed action vector. This in turn leads to the next state that falls within the range stochastically governed by the action parameters. A fixed action operator is always associated with a fitness value given by (6). By this token, an action operator can be applied anywhere in the state space with an associated fitness value evaluated from the knowledge of the environment gathered so far in terms of the experience particles held in memory (i.e. the training set $\Omega$). Further, if the control policy is defined through a fitness value prediction, such as the softmax policy in (10), the action operator also represents an abstraction of a policy. This is due to the fact that the action operator encodes both a particular state transition and its fitness value once the state and action vector are known. Thus, by fixing a state, all (parameterized) action choices can be compared in a similar way through their corresponding fitness value estimates. The result of the fitness value prediction is governed by the property of the reinforcement field, which is established by all the known experience particles distributed over the state space. In this way the reinforcement field connects with the action operator, which, in turn, determines the continuous process of the policy inference. A high-level policy inference in a reinforcement field is illustrated in Fig. 3. The center circle represents the goal area. Also marked in the center circle are 12 possible motions, each with an allowable range for action variations. The outer three pairs of the shaded circles enclose the experience particles (in dots) holding control policies suggested by the associated augmented states. The new query points are marked by crosses. As long as the query point (e.g., an unexplored state) is sufficiently close to the states of the neighboring particles (i.e., within the scope of the shaded circles), the resolved action is expected to be similar (in terms of action parameters) to those in the particles. The exact inference is given in Section IV.

To sum up, a reinforcement field shares the following properties analogous to other physical fields: i) a source of origin ii) a spatial intensity and iii) a governing law that acts on particles in the field. The field source originates from experience particles, each of which embeds a control policy that generalizes into the neighboring states through the kernel. The intensity of an experience particle is represented by the fitness value evaluated from the combined effect of all the other particles in proportion to their degree of correlation and their intensities (i.e., their fitness values). Upon any state transition, a new particle is created that encapsulates the value of the state vector, action vector, and the mapping to a fitness value. The new particle joins the rest of the population to form an updated reinforcement field that subsequently determines the fitness value of yet another new particle.

## IV. Policy inference via Particle reinforcement

Because the policy learning in the reinforcement field depends on experience particles, a sample update rule is required so that the memory will not exceed a given bound. Several aspects play into the role of constructing a proper mechanism for the particle evolution. First, the new particle retained in the memory needs to reflect the latest dynamics of the system so that the policy generalized from these particles will be as accurate as possible. This can also be explained by noting that the policy is driven by the following two continuous processes in parallel: i) the kernel hyperparameter tuning by ARD and ii) the fitness value prediction by GPR.

Recall from (8) that the ARD procedure identifies the hyperparameters in a manner that trades off between data fit and model complexity, leading to the best explanation of the data (from the perspective of maximizing the marginal likelihood). Thus, a periodic update using ARD is a key mechanism for the reinforcement field to keep up with the latest utility signals from the underlying control learner (to be discussed further). The process ii) above is just the flip side of the same coin in that the current fitness values referenced by the experience particles in memory effectively assimilate the update history of the utility estimates from the underlying control learner. Thus, the prediction for a new augmented state will only be as good as the hyperparameters.

On the other hand, since the goal of policy learning is to steer the agent towards a state sequence leading to a maximum utility (as a global constraint), it is imperative for the agent to retain the particles aligned with that objective. This is to say that in each step of the decision process, the action operator ideally should trigger the state transition (through the local constrain enforced by the action parameters) to follow the path in the direction of increasing the fitness value.

Note that the policy inferred from the action operator at a given augmented state $(\mathbf{x}_s, \mathbf{x}_a)$ is locally applicable to all the neighboring states so long as they are within the scope of a given correlation degree measured by the kernel. This is a property of policy inference in the reinforcement field in which any given two particles are considered similar as long as the combined effect of the state and action ends up being similar in terms of their fitness (Fig. 3). Yet from another perspective, this policy generalization property is indeed desirable because the state transition given a fixed state and action will likely vary over time, depending on the action parameters as a random process. The random variations of the action parameters encode the uncertainty in the action dynamics, which can either be interpreted as the uncertainty inherent in the action performance or the uncertainty resulted from the surrounding environment.

We will discuss more of the equivalency of stochastic effects inherent within the (observed) state and the action in addition to their implications in Section B. Moreover, depending on the exploration strategy underlying the policy search, the agent can choose to always follow a stochastic policy or a greedy policy after sufficient exploration of the state space. In the empirical study, we will consider a softmax exploration strategy in (9) that gradually reduces to an ε-greedy policy.

### A. Experience Association and Particle Evolution

For the agent to have a mechanism to "interpolate" the best policy given the experience particles, we shall now introduce a central operation for the particle update – *experience association*. Additionally, two attendant definitions are given: i) hypothetical state and ii) particle polarity.

**Definition 5 (Hypothetical State)** *A hypothetical state $s_h^+$ is an augmented state formulated by the agent at a given state $s \in S$ ($s = \mathbf{x}_s$) replicating the primitive action $a \in A$ ($a = \mathbf{x}_a$) from an experience particle $\omega \in \Omega$ to give $s_h^+ = (\mathbf{x}_s, \mathbf{x}_a)$.*

**Definition 6 (Particle Polarity)** *Positively polarized particles, or positive particles, are those associated with a non-negative*

*temporal difference (i.e., TD ≥ 0) whereas negatively polarized particles, or negative particles, are associated with a negative temporal difference (i.e., TD < 0).*

The operation of *experience association* is inspired by a basic learning principle: if some decisions had led to a desired outcome in the past, then the same strategy can be readapted for similar situations in the future, recreating a successful experience. The context of this new situation is provided by the state and partly by the action operator as well, as they represent mutually interactive entities, while the desirability of the action is quantified by the fitness value. An example can be observed in Fig. 3 where the ideal action at the query points A, B and C are aligned with the control policy embedded in their neighboring particles. Whether two particles are (sufficiently) correlated is again determined by the kernel function.

Formally speaking, the experience association operation is performed by the agent at a state $s = \mathbf{x}_s$ and tries to access an instance of the past memory, say, $s_p^+ = (\mathbf{x}'_s, \mathbf{x}_a)$, encapsulated in a particular experience particle, $\omega \in \Omega$, which also carries a fitness value. The agent subsequently "extracts" the action vector $\mathbf{x}_a$ from the target memory $s_p^+$ – and combine it with its current state $\mathbf{x}_s$ to form a new, hypothetical state $s_h^+ = (\mathbf{x}_s, \mathbf{x}_a)$ (Definition 5). The resulting hypothesis $s_h^+$ is then compared, with a threshold, with the target particle $s_p^+$ using the given kernel function $k(\cdot,\cdot)$. That is, by evaluating $k(s_h^+, s_p^+)$, where $s_h^+ = (\mathbf{x}_s, \mathbf{x}_a)$ and $s_p^+ = (\mathbf{x}'_s, \mathbf{x}_a)$, if $k(s_h^+, s_p^+)$ goes above a pre-specified threshold $\tau$, i.e., $k(s_h^+, s_p^+) \geq \tau$, then $s_h^+$ is said to be (strongly-)correlated to the past memory state $s_p^+$; on the contrary, if $k(s_h^+, s_p^+) < \tau$, then, $s_h^+$ is considered as not sufficiently correlated to this past memory state $s_p^+$. The implication of viewing a particle generated in the past as being sufficiently correlated is that, if the agent were to take this hypothetical action from the past and replicate it in its current state, then it is expected to observe a similar fitness value given that the correlation takes into account both the state (i.e. the context) and the action. In this manner, the threshold $\tau$ effectively determines the size of the shaded circles in Fig. 3, meaning that if we set a small threshold such that it is relatively easy for the agent to view a past instance as being correlated, then it will end up having a large sample size of candidate particles (contexts) from which their corresponding actions can be extracted and taken while remaining correlated to its current state (situation). A visual example can be found again in Fig. 3, where the query point A along with its neighboring states all share the same 2-o'clock action as being the optimal choice. Conversely, if we set a large threshold, then the sample size of such correlated particles would be relatively small if they exist. By applying experience association, effectively, the agent is given a higher preference for the policy embedded in a neighboring particle than those further apart.

To enable the agent to tell apart desirable actions from undesirable ones, this version of GRL adopts a dichotomy by labeling particles in terms of the temporal difference (TD) obtained from the base control learner (Definition 6). Recall from (2) that a non-negative TD (i.e., $R(s, a) + \gamma Q(s', a') \geq Q(s, a)$) implies an improvement in the utility estimate and the corresponding action is in general preferable to those with negative TDs. As such, the particle polarity can be used as a guidance to evolve experience particles in the desirable direction. Specifically, if we wish for the action operator to resolve into a state transition along the direction that increases the fitness value, the existing particles representing past memory are to be replaced by the new particles with relatively higher fitness values.

---

**Algorithm 1: MemoryUpdate**($p$, $TD$, $k$, $\tau$, $\Omega$) return $\Omega_{new}$

**Input:**
    $p$: A candidate particle comprising an augmented state and (optionally) its associated fitness value
    $k$: Kernel function $k(\cdot,\cdot)$
    $\tau$: Threshold of correlation (lower bound)
    $\Omega$: The working memory holding experience particles
    $TD$: Temporal difference associated with the new particle $p$
        //e.g., for SARSA, $TD(s, a) = R(s, a) + \gamma Q(s', a') - Q(s, a)$

1. // Rank all the correlated and polarity-aligned particles $\omega \in \Omega$
  // in the descending order of degrees of correlation to the
  // new particle $p$ using a (max-)priority queue
  **set** $maxq$ to an instance of max queue
  **for** $\omega$ in $\Omega$: //for each particle in memory
    **if** $(k(p, \omega) \geq \tau)$ **and** $(polarity(\omega) == polarity(p))$ **then**
      $maxq.push(\omega)$ //push $\omega$ into the max queue
    **end if**
  **end for**

2. **while not** $maxq.empty()$:
    $\omega = maxq.pop()$ // take the most correlated particle w.r.t. $p$
    **if** $TD \geq 0$ **then**
      // $TD(p)$ is positive, apply the rule for *positive particles*
      **replace** $\omega$ by $p$ iff $\omega$ has a smaller fitness value
    **else** // $TD < 0$     // $TD(p)$ is negative
      // $TD(p)$ is negative, apply the rule for *negative particles*
      **replace** $\omega$ by $p$ iff $\omega$ has a larger fitness value
    **end if**
    **if** a match is found, **then**
      **break** // update complete, jump out of the loop
    **end if**
  **end while**

3. **return** $\Omega_{new}$   // $\Omega_{new} \neq \Omega$ if replacement occurred

---

However, care must be taken in such replacements. The new experience particle subsumes an old one only when they are sufficiently correlated. For a particle update to occur, first, the polarity must be aligned: particles referencing similar states (e.g., nearby positions in a navigation task) but with an opposite polarity are indicative of a high degree of dissimilarity in the corresponding actions. This gives rise to the memory update rule in Algorithm 1, referred to as the particle reinforcement (sampling) algorithm henceforth, because such update rule promotes a "stronger" reinforcement field by preserving particles leading to higher fitness values.

Algorithm 1 is organized in units of logical steps (1 through 3) instead of the conventional line numbers for ease of exposition. The associated subroutine for particle reinforcement is mnemonically called *MemoryUpdate* for its apparent function of organizing the agent's memory, assimilating new learning experience; we will reuse this subroutine to develop

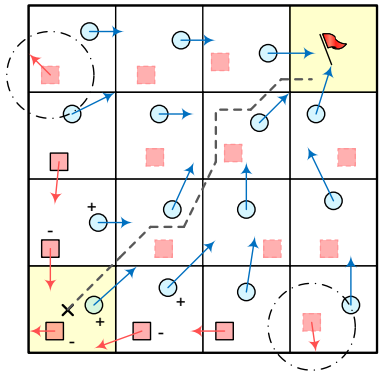

Fig. 4. A 2D navigation domain where the agent seeks an optimal path to retrieve the flag following the shortest route. Positive particles are marked in blue that locally serve as the guiding signal, leading the agent to traverse along an optimal path, whereas negative particles are marked in red that serve as the constraints, informing the agent which actions (and hence directions) to avoid at its current position.

further algorithms in later sections.

Step 1 in particle reinforcement begins by ranking existing particles according to their degrees of correlation and polarities with respect to the input particle $p$: if the particles in memory $\{\omega\} \subseteq \Omega$ are sufficiently correlated, by the condition $k(p, \omega) \geq \tau$), and are polarity-aligned with $p$, then push qualified particles into a max-priority queue so that they can be later retrieved (at Step 2) in descending order of degrees of correlation. Note that the input $p$ can be an new experience particle or simply a hypothetical state $s_h^+$ without a fitness value.

The replacement rule (Step 2) in *MemoryUpdate* assumes that initial fitness values are not overestimated. Only under such condition will the positive particles continue to increase their signal strengths (i.e., fitness values) throughout the learning cycles. The update rule for positive particles is logically opposite to the rule for negative particles: the TD values for positive particles are expected to increase over time as the policy learning proceeds, whereas the negative particles have decreasing TD values. The net result for this polarity-aligning update rule is to drive the positive particles toward a progressively improved policy while the negative particles continue to serve as counterexamples, or constraints, throughout the learning cycle. The iteration for the particle replacement (Step 2) terminates as soon as a match is found; otherwise, no update will occur if no match is found among the current population of particles in memory.

Note that, by Definition 3, an experience particle holds only two pieces of information: an augmented state (a vector) and a fitness degree (a real number). In the particle reinforcement algorithm, however, we also need access to the polarity of the queried particle $\omega \in \Omega$ in order to check for their alignments with the input $p$. Note also that, in practice, the TD value associated with the input, denoted $TD(p)$, is pre-computed externally before passing as an argument to the particle reinforcement subroutine *MemoryUpdate()*. This is because, under the two-tier approach for fitness value estimation (Fig. 2), Q-value estimates at the base layer (running a TD learning algorithm in MDP) is passed to the GPR layer as "labels" for fitness estimation, which is framed as a regression problem.

$TD(p)$ depends on the Q-value since it really comes from part of a temporal-different learning update rule, which, in the case of SARSA, can be obtained by (2):

$$[R(s,a) + \gamma Q(s', a')] - R(s, a).$$

From the engineering standpoint, we may wish to tag additional meta data to a particle such as its associated TD value (which carries the polarity information), among other quantities, but conceptually, it is helpful to simply think of a particle as a vector paired with a functional response.

When the size of the training set $\Omega$ is large, a global search for a matched particle can be expensive. To reduce the computational cost, Step 1 and 2 in Algorithm 1 can be extended to incorporate more efficient search strategy in practice such as maintaining particles in a K-D tree or using approximate search via locality sensitive hashing (LSH) [43]. To further allow for the RL agent to explore the state space efficiently while reducing the search cost in organizing its memory, we can consider subdividing the state space into sub-areas $\{S^{(k)}\}$ such that each sub-area $S^{(k)} \subseteq S$ references a fixed number of positive and negative particles, where $S^{(k)}$ can be either a simple grid partition or a subgoal area in line with the HRL method [13]. In this manner, the expensive global search can be reduced to a simpler local search whenever the agent updates its memory through particle reinforcement. Specifically, when the agent at state $s$ (with value $\mathbf{x}_s$) creates a new particle associated with a hypothetical state $s_h^+ = (\mathbf{x}_s, \mathbf{x}_a)$ by taking an action in parametric form $\mathbf{x}_a$, the value $\mathbf{x}_s$ can be used to identify its associated partition, say, $S^{(k)}$, via a space-partitioning data structure such as a KD-tree. Once the state partition is located, we can assume (with a caveat) that particles within the same state partition, i.e., $\{(\mathbf{x}_s, \mathbf{x}_a)_j^{(k)}\}$ with $(\mathbf{x}_s)_j^{(k)}$ falling into the boundary of $S^{(k)}$, will have a relatively greater chance to match the target hypothetical state $s_h^+$. Note that we have left out the particle's fitness value for simplicity. We will refer to the partition $S^{(k)}$ associated with the agent's current state as the current partition. Particles associated with the other partitions are generally considered as less correlated to those in the current partition with one caveat: the agent's current state can be near the boundary of $S^{(k)}$ where potentially other particles can exist in the adjacent partition(s) (and hence these "foreign" particles can be strongly correlated to $s_h^+$). We will address such "boundary conditions" in future versions of GRL.

Perhaps the particle reinforcement algorithm can be best visualized through a robot navigation domain as illustrated in Fig. 4, where the agent aims to travel from the start position to the goal area following a shortest path. The start position is marked by a cross in the lower left square and the goal area is enclosed by the upper right square with a flag. The agent has the freedom to move in all directions within the boundary of the continuous state space (i.e., borders marked by thick lines). The allowable actions include the 12 clock-direction movements parameterized by a radius and an angle increment $(\Delta r, \Delta \theta)$ adopted from the simple parametric action model for 2D navigation depicted earlier in Fig. 1 and 3. The state space is partitioned into a 4-by-4 grid for the purpose of estimating

**Algorithm 2: RF-SARSA($k, M, T, \Omega_0, \tau$)**

**Input:**
- $k$: Kernel function $k(\cdot,\cdot)$ with hyperparemters $\boldsymbol{\theta}$ (e.g. (4))
- $M$: Parametric action model with $n$ primitive actions:
  $\{a^{(i)} \in A \mid i = 1 \sim n\}$ along with the action mapping
  $f_{A^+}: A \to A^+$ and the inverse mapping $f_A: A^+ \to A$
- $T$: Cycle for ARD procedure, which runs once every $T$ state transitions
- $\Omega_0$: Initial source of particles (or simply an empty set)
- $\tau$: Threshold of correlation (the lower bound)

1. **Run ARD** to estimate the initial hyperparameters $\boldsymbol{\theta}$ using $\Omega_0$, or simply start with an initial prior $\boldsymbol{\theta}_0$
2. **Initialize** $Q(s, a^{(i)})$ arbitrarily
3. **Initialize** particle set (i.e., training set for GPR): $\Omega \leftarrow \Omega_0$

**Repeat** (for each episode)

4. **Initialize** $s$.
5. Given a state $s$, compute $Q^+$ for each $a^{(i)} \in A$; i.e.,
   **foreach** $a^{(i)} \in A$:
   5.1 Compute the augmented state $s^+$ by resolving $a^{(i)}$ into its vector form $\mathbf{x}_a^{(i)}$ according to $M$, along with the state vector $\mathbf{x}_s$ to get
   $s^+ = (\mathbf{x}_s, \mathbf{x}_a^{(i)})$
   5.2 Predict $Q^+(s^+)$ using GPR (see Eq. (3) and (6))
6. **Choose** $a^{(i)}$ from $s$ following the current policy: $\pi(s, a^{(i)}) = \pi[Q^+]$ (e.g., softmax policy (10) using $Q^+$ value from Step 5

**Repeat** (for each step of episode):

7. **Run ARD** to update $\boldsymbol{\theta}$ every $T$ steps (state transitions)
   7.1 (Optional) $T \leftarrow T + f(T)$ // Gradually increase $T$
8. From $s$ **take action** $a^{(i)}$ and observe $r$ and $s'$
9. **Choose** $a^{(j)}$ from $s'$ using $\pi[Q^+]$, where $Q^+$ is again estimated through the parametric action model and the GPR layer as specified in Step 5, which can be organized into a reusable subroutine
10. **Update** $Q(s, a^{(i)})$ using (2) with the observed reward $r$ from Step 8    // See Section II.A
    10.1 **Save** the new Q-value estimate for later use
    $q_i \leftarrow Q(s, a^{(i)})$
    10.2 **Save** the temporal difference from (2), i.e.,
    $TD(s, a^{(i)}) = [R(s, a^{(i)}) + \gamma Q(s', a^{(j)})] - Q(s, a^{(i)})$
    $t_i \leftarrow TD(s, a^{(i)})$
11. $s \leftarrow s'; a^{(i)} \leftarrow a^{(j)}$

//Now, update the working memory …

12. **Propagate** the value estimate $q_i$ from Step 10 to the GPR layer, and **generate** a new particle $\omega$ that references both the augmented state $(\mathbf{x}_s, \mathbf{x}_a^{(i)})$, where $\mathbf{x}_a^{(i)}$ is the parametric form for $a^{(i)}$, and the fitness estimate $q_i$, i.e., $\omega = ((\mathbf{x}_s, \mathbf{x}_a^{(i)}), q_i)$
13. **Run particle reinforcement** via **MemoryUpdate**($\omega, t_i, k, \tau, \Omega$) to update working memory $\Omega$, assimilating new particle $\omega$
    // $t_i = TD(s, a^{(i)}) = TD(\omega)$ is precomputed at Step 10

**Until** $s$ is terminal or max number of episodes reached

---

utility values associated with each state subarea as an approximation to a continuous-state MDP.

Now let us consider SARSA as the base control learner that obtains accumulated rewards as guiding signals for estimating fitness values, using GPR, for each particle (see Fig. 2 for a schematic view of this two-tier value estimation approach). In Fig. 4, each state partition is configured to reference only two particles with one being positive and the other negative. Positive particles are depicted by small blue circles, each of which points to a relatively preferable direction to move. By contrast, negative particles are represented by the small red squares. To keep the figure uncluttered, only selected negative particles are marked with arrows indicating their navigational policies.

Recall that an action operator effectively combines both the state and action information (as represented by the augmented state) and is always associated with a fitness value after all the parameters are fixed. Therefore, we can associate each particle as carrying a "functional signal" that maps a control policy (indicated by the underlying state) to a fitness value (indicating the signal strength) in the reinforcement field. More specifically, through the property of the kernel given by (12) and the GPR prediction in (6), all the red and blue particles combined establish a particular reinforcement field, a vector field of functions, representing a policy generalization mechanism surrounding the neighboring state of the particles.

Note that two conditions limit the range of the motion given an action choice: i) constraints imposed by the action parameters (e.g., feasible range) and ii) the exterior boundary of the entire state space. To facilitate the construction of a new SARSA algorithm that assimilates the notion of the particle reinforcement, it helps to simplify the simulation physics for robot navigation as given by Fig. 4. That is, we will further assume that, as the robot navigates, if the action operator resolves into a position and a motion that leads outside of the spatial boundary, then the action will only be executed up to the boundary but not beyond. With this physics assumption, the agent will move as far as it hits the border and then will remain there until the next decision step. This is illustrated by the dotted circle enclosing two example negative particles (in red squares), each of which embeds a motion that eventually hits the border.

Illustrated in Algorithm 2 is an example RF-SASAR, where SARSA, as the baseline TD-learning algorithm, is applied within the **r**einforcement **f**ield (RF). As before, the algorithm is organized in units of logical steps (1 through 13) instead of the line numbers for ease of exposition.

In Step 1, we run the very first ARD procedure (see Eq. (8) and (9)) to estimate the initial hyperparameters $\boldsymbol{\theta}_0$ associated with the input kernel function $k(\cdot,\cdot)$, which is expected to change over time as new evidence, or experience particles, are collected in the memory $\Omega$. At this stage, we could either assume that there exist seed particles such that $\boldsymbol{\theta}_0$ can be bootstrapped or simply set it to a reasonable initial value. As the RF-SARSA proceeds with new experience particles accumulated, the same ARD procedure is to be executed periodically every $T$ state transitions, as specified in Step 7, to re-estimate the kernel hyperparameters, which continues to factor in new TD estimates (and evolving system dynamics).

The step 2, 4, 6, 8, 9, 10, 11, in that order, are reminiscent of those in standard SARSA. The presence of the parametric-action model can be traced in multiple steps: Step 5, where primitive actions are mapped to their parametric forms; Step 6 and 9 through the policy function $\pi[Q^+]$, where the fitness function $Q^+$ takes on the augment state, representing the generalized state-action pair; Step 12, where experience particles are formed and the reinforcement field is established through these particles maintained in the memory $\Omega$; and Step 13, where particles are updated and evolved by incorporating new evidence through TD signals and experience/memory association. Likewise, in Step 1 and 7, where kernel hyperparameters are updated, parametric forms of actions are retained through the fitness function $Q^+$.

Zooming in more closely, we see that Step 2 is the initialization step for the Q-values at the base level (i.e., lower tier in the hierarchical learning architecture depicted in Fig. 2) in contrast to the GPR layer at Step 12 (i.e., upper tier), where the training examples for the fitness function approximator are in the form of a function data set comprising experience particles. Step 2 initializes a Q-Table if both the state $s \in S$ and the primitive action $a^{(i)} \in A$ are discrete, or discretized if they are continuous. Step 2 can be extended to incorporate the continuous MDP by sampling a subset of states $\{s^{(i)} \in S \mid i = 1 \sim m\}$ so that a finite sample of size $m$ is obtained. We can then approximate the value function $Q(\cdot)$, at these sampled states, by treating it as a regression problem, much the same way as how the fitness value function $Q^+(\cdot)$ is represented by the GPR. It is certainly conceivable that $a^{(i)}$ is also by itself continuous like its parametric form $\mathbf{x}_a^{(i)}$. This corresponds to the case in Fig. 1, where, instead of considering a finite set of clock directions, we have infinitesimal directions, each of which is parameterized by the radius- and angle-specific error, as a random vector, to account for the uncertainty associated with the erratic behaviors of actions when carried out in a real-world system. With the continuous MDP, we end up solving two regression problems, with one being at the level of primitive actions and the other at the level of parametric actions. To focus on the SARSA variant in the reinforcement field, however, Algorithm 2 assumes that both the state and the primitive action are either discrete or can be discretized, giving rise to a simpler and cleaner form of the RF-SARSA.

Step 3 can be thought of as an initialization step for the GPR layer in preparation for the future ARD procedures within the loop of state-action-reward-state-action (abbreviated as the "sarsa" loop) spanning from Step 7 through 11 – plus the memory-specific operations in Step 12 and 13. Alternatively, we could also merge this step into Step 1 as its precondition by assuming that some seed particles ($\Omega_0$) exist a priori and are assigned to the memory $\Omega$, whereby initial kernel hyperparameters are estimated.

As in standard SARSA, Step 4 initializes the starting state, where the agent selects the first action according to the initial policy $\pi$ (Step 6), followed by entering the sarsa loop. Step 5 coupled with Step 6 are the key step, through which the standard SARSA gets transformed into its RF counterpart. In RF-SARSA, we need to consider how a chosen primitive action $a^{(i)}$ is mapped to its parametric form $\mathbf{x}_a^{(i)}$ via the primitive-to-parametric-action map $f_{A^+}: A \to A^+$ and subsequently estimate the fitness value of the corresponding augmented state $s^+ = (\mathbf{x}_s, \mathbf{x}_a^{(i)})$ by computing $Q^+(s^+)$ through GPR. Recall from Section II.E that the map $f_{A^+}$ is task-dependent and is estimated externally (see Section II.B for an example) prior to running the RF-SARSA. It is also possible to interpret $f_{A^+}$ as a mapping from the primitive action to an embedding space (i.e., action embeddings), where action parameters become learnable parameters themselves (like kernel hyperparameters) such that they can be optimized with respect to a chosen optimization objective either locally (e.g., approximating fitness values) or globally (e.g., achieving maximal accumulated reward). For simplicity, we will consider $f_{A^+}$ as given a priori through the input parameter $M$ and leave the learning of $f_{A^+}$ in other works.

Within the sarsa loop (Step 7 through 13), the ARD is run periodically at Step 7 so that the kernel hyperparameters can be progressively updated to capture the most recent memory $\Omega$, in which new particles are collected at Step 12 and subsequently replace the old particles (that are sufficiently correlated) at Step 13 via the particle reinforcement subroutine. Step 8 through 11 are fundamentally similar to the corresponding steps in standard SARSA except for the RF-specific additions: in Step 9, we estimate the fitness value $Q^+$ for candidate (generalized) state-action pairs using $f_{A^+}$ and GPR such that the (desired) action, $a^{(j)}$ with its parametric form $\mathbf{x}_a^{(i)}$, can be sampled according to $\pi[Q^+]$ that gives proportionally higher probability mass to the actions leading to higher fitness values; in Step 10, the Q-value for the resulting state and (primitive) action pair $Q(s, a^{(i)})$, or $q_i$, is cached and later passed on to the GPR layer at Step 12 to form an experience particle $\omega$, or $\left((\mathbf{x}_s, \mathbf{x}_a^{(i)}), q_i\right)$, while the corresponding TD signal $t_i$, or $TD(s, a^{(i)})$, is passed to the particle reinforcement subroutine at Step 13.

Last but not least, notice that the very same kernel function $k(\cdot,\cdot)$, as an input to the RF-SARSA, is used in multiple steps: in Step 5, $k(\cdot,\cdot)$ serves as the covariance function for the GPR (and as the basis for computing covariance matrix $K$ internally) to make fitness value predictions; in Step 6, $k(\cdot,\cdot)$ indirectly influences the probability of choosing a (primitive) action, $\pi(s, a^{(i)})$, given that the policy $\pi$ is a function of $Q^+(\cdot)$, which is modeled by the GP; in Step 13, where particle reinforcement occurs by calling the subroutine *MemoryUpdate*, the same kernel $k(\cdot,\cdot)$ is used as the basis for experience association whereby old particles in the memory $\Omega$ are systematically replaced by new particles, which carry the most recent TD information from Step 10. Later in the CDLA, we will see that $k(\cdot,\cdot)$ can also be used to group experience particles by viewing the particles in memory $\Omega$ as a graph, which represents the agent's ongoing learning experience, and solving for the graph partition problem as such allows for the agent to form levels of abstraction and learns an abstract-level control policy.

### B. Soft State Transition

Action operators, similar in a way to the primitive action set in standard RL, are chosen to represent only the local dynamics in the state space. Under the MDP formulation, executing a primitive action triggers a state transition in which the action can be considered as one of the variables in addition to those representing the state. With action operators, the concept of state transitions is extended further to the transition between pairs of correlated augmented states, each of which combines the state vector and the action vector with their vector components (or feature values) determined only upon their merge. As mentioned earlier, the uncertainty captured by action parameters can be explained away using an equivalent model over stochastic observations of the state.

The top of Fig. 5 conceptualizes the dual property of the uncertainty in the action and in the corresponding start state adapted from the 2D navigation domain. Like Fig. 3, a dot (i.e., a state) combined with an arrow (i.e., an action) represents a local policy referenced by an experience particle. The uncertainty modeled by action parameters result in two of the possible motions on the left while equivalently on the right indicate two deterministic actions (i.e., the 3-o'clock action with uncertainty removed via averaging) applied to two nearby states that lead to the same successor states. The bottom half of Fig. 5 illustrates 3 trajectories (all starting from the leftmost circle) consisting of "microscopic" state transitions, all of which lead to the same general region as enclosed by the rightmost circle ($s_5$) with a flag within the boundary. On a larger scale, however, all the trajectories approximately follow the same route in terms of the sequence of the 5 successive circles from $s_1$ to $s_5$. Each macroscopic state as such, as represented by the circle, has a non-zero volume, and is referred to as a *soft state* in this paper, as opposed to a concrete, microscopic state, being essentially a point in the state space. Note that we will reuse the same notation for the micro-state (i.e., $s \in S$ as defined via MDP) and the macro-state (or soft state) to avoid cluttering of symbols when there is no ambiguity.

The state transition by virtue of correlation under experience association effectively relaxes the start and end points for each transition; namely, as long as the state and action vector are jointly correlated in terms of the kernel measure (i.e., $k(\mathbf{x}, \mathbf{x}') \geq \tau$), their combined effect will lead to approximately the same state region enclosing yet another set of highly correlated augmented states with correlated fitness values. In other words, a *soft state transition* as such bridges any two sets of highly correlated states with their volumes effectively controlled by the threshold parameter $\tau$ in the kernel-induced feature space.

The soft state boundary will remain identical in the entire state space for a stationary kernel that is translation invariant (e.g., SE kernel in (4), Matérn class of kernels [8], etc.); however, the boundary will vary in the state space with a non-stationary kernel such as the dot product kernel:

$$k(\mathbf{x}, \mathbf{x}') = \sigma_0^2 + \mathbf{x}^T \Sigma \mathbf{x}',$$

where $\Sigma$ represents a covariance matrix for the input vector components. In Fig. 5, the 3 hollow dots from within the soft state $s_3$ reference 3 out of 9 possible concrete state transitions to the corresponding hollow dots in the soft state $s_4$; however, the 9 state transitions can all be captured, on a larger scale, by the same one-step soft state transition (albeit with less accuracy).

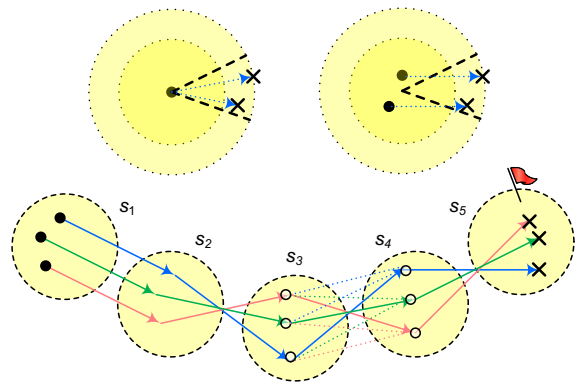

Fig. 5. Soft state transitions: the top two concentric circles contrast two equivalent one-step transitions with uncertainty captured by action parameters (left) and observed states (right), respectively. Enclosed within the 5 "soft states" below in circles indicate 3 out of the $3^4$ possible "microscopic" transitions starting from the leftmost soft state to the rightmost soft state with a flag within its boundary, all of which correspond to a single "macroscopic" soft state transition.

As an exploratory subject for developing future generations of GRL agents, it is helpful to further define the soft state transition in a strict sense by requiring reachability for particles across the soft state boundary. The weaker assumption given earlier only requires that the constituent particles in one soft state being correlated, or coupled, with those from another soft state, forming "weak bonds." On the other hand, if particles across the soft state boundary are mutually reachable through any subset of (feasible) actions (i.e., $a \in A$ assuming values $\mathbf{x}_a \in A^+$), then the actual state transition, as the name suggests, is possible, and therefore, we would describe these particles across the boundary as forming "strong bonds." Each individual strong-bond soft state transition is essentially the same as the state transition defined in the standard MDP formulation. As a corollary, if the particles in one soft state $s_i$ are both coupled and mutually reachable with those from a different soft state $s_j$ ($j \neq i$), then the same particles, by definition, must also be both coupled and mutually reachable with the other particles in the same soft state, given that a soft state is effectively a cluster of correlated microstates under experience association defined by the kernel-based criterion. We will leave the detailed treatment of the soft state transition to future research.

### C. The POMDP Perspective

The soft state transition through experience association can also be achieved via a probabilistic model that identifies an unexplored yet correlated state by drawing an analogy to the belief state formulation in POMDP. Recall from Section II that an MDP is characterized by the 4-tuple $\langle S, A, T, R \rangle$. For an environment where states are only partially observable, two additional elements are incorporated into the representation to capture observations of states, giving rise to the POMDP extension with the tuple $\langle S, A, T, R, O, Z \rangle$. In particular, $O$ represents the observation space and $Z$ represents an observation function in terms of the conditional probability $Z(s', a, o') = p(o_t = o' | s_t = s', a_{t-1} = a)$ that is computed along with the transition probability $T(s, a, s') = p(s_t = s'|s_{t-1} = s, a_{t-1} = a)$ as the agent traverses the environment. While $T$ specifies the probability of transitioning from the state $s$ to $s'$ by taking the action $a$, correspondingly $Z$ specifies the probability of receiving an observation $o$ at $s'$ in response to the same state transition. Due to the issue of partial accessibility of the state, the transition function $T$ alone can no longer be used to specify the agent's worldview regarding the state transition. Instead, the agent maintains a probability distribution over the (true) state given an often-incomplete measure of the state; namely, an observation. The observation model $Z$ combined with $T$ is then used to draw a conclusion of the most probable next states in terms of a conditional probability distribution, which gives rise to the notion of a belief state.

A belief state is typically modeled by a conditional probability density of the current state given the history, i.e.

$$b_t(s') = p(s_t = s' | I_t), \tag{14}$$

where $I_t$ denotes the history of the observed states up to the time step $t$. In particular,

$$I_t = (o_0, o_1, ..., o_t, a_0, a_1, ... a_{t-1}), t = 1, 2, ...$$
$$I_0 = o_0$$

By applying Bayes' law and subsequently the Markovian property on the observations and the states, (14) can be greatly reduced to:

$$b(s_t = s') \propto p(o_t | s_t, a_{t-1}) \int_S p(s_t | s_{t-1}, a_{t-1}) b_{t-1}(s_{t-1} = s) ds_{t-1} \tag{15}$$

Note that (15) omits the normalization factor in the denominator for simplicity and, in contrast to (14), does not depend on the past observations $I_{t-1}$ including the past action sequence. By simple comparison, the RHS of (15) amounts to $Z(o', a, s') \int_S T(s, a, s') b(s) ds$ using the definitions the state transition and observation functions $T$ and $Z$.

Inferring the current state using the belief-state model is analogous to using experience association over the experience particles represented through the functional data set $\Omega_m : \{(\mathbf{x}_1, q_1), ..., (\mathbf{x}_m, q_m)\}$. In particular, by comparing (6) and (12), the weights $\{\alpha_i\}$ in (12) can be re-expressed in a vector form: $\boldsymbol{\alpha} = K(X,X)^{-1}\mathbf{q}$, making it explicit that $Q^+$ is a linear combination of kernels centered on the augmented states. Alternatively, by setting $h(\mathbf{x}_*) = K(X,X)^{-1}\mathbf{k}(\mathbf{x}_*, X)$, the fitness value function (12) can be expressed explicitly as a linear combination of other historical signals (i.e., the fitness values associated with the other particles in the memory):

$$Q^+(\mathbf{x}_*) = \mathbf{k}(\mathbf{x}_*, X)^T K(X,X)^{-1} \mathbf{q} = h(\mathbf{x}_*)^T \mathbf{q}, \tag{16}$$

where $h(\mathbf{x}_*)$ is also known as the *weight function* in the context of *kernel regression* [30].

From the form of $h(\mathbf{x}_*)$, it is not difficult to see that its value depends on the data and but not on the prediction; that is, $h(\mathbf{x}_*)$, is driven only by the binding of the state and the action expressed through the augmented states (the data portion of the particle) (i.e. $\{\mathbf{x}_i\} = \{(\mathbf{x}_s, \mathbf{x}_a)_i\}$) but not on their fitness values $\{q_i\}$ (the signal part of the particle). If the particles in the **x**-space become so dense that $h(\mathbf{x}_*)$ approaches a continuous curve, then $h(\mathbf{x}_*)$ can be analyzed in terms of the power spectrum of $k(\cdot, \cdot)$ and is also referred to as the *equivalent kernel* (EK) by an analogy to *kernel smoothing* [29], [30]. The EK generally exhibit a localization property with its value higher for the neighboring points with respect to the test point and decaying with distance. The localization property becomes more prominent as the number of training points increase. For instance, the EK for the SE kernel in (5) can be approximated by a sinc function [29]. By (16) and the continuity assumption of the augmented state space (i.e. the **x**-space), $h(\mathbf{x}_*)$ as the EK effectively encodes the correlation degrees of the queried state $\mathbf{x}_*$ with respect to all the other particles in the memory where $\mathbf{x}_*$ effectively serves as the "center" for the evaluation. This again justifies that the fitness value prediction by (16) will weigh heavier on more correlated pieces of evidence.

Although the localization property of the EK suggests its use as a kernel density estimator for building a probabilistic model for comparing experience particles, it is made complicated by the fact that EK generally has a higher spatial frequency than the original kernel $k(\cdot, \mathbf{x}_*)$ for points away from the center. That is, while $k(\cdot, \mathbf{x}_*)$ is always non-negative, the curve of the EK has zero-crossings similar to the side lobe of a sinc function. Nevertheless, in the case where $k(\cdot, \cdot)$ is also a localized kernel (e.g., SE kernel), an adaptation to the belief state model in (15) can then be achieved by reusing the associated kernel expressions in $h(\mathbf{x}_*)$ to define the weight:

$$w_i = \frac{k(\mathbf{x}_*, \mathbf{x}_i)}{\sum_{i=1}^m k(\mathbf{x}_*, \mathbf{x}_i)} \tag{17}$$

By (17), the fitness value function can alternatively be predicted through a linear combination of these weights: $\hat{Q}^+(\mathbf{x}_*) = \sum_{i=1}^m w_i q_i$. This is also known as the Nadaraya-Watson model [8, 30]. Because of the constraint: $\sum_i w_i = 1$, each weight $w_i$ can be endowed with a probabilistic semantic as being the probability of the test point $\mathbf{x}_*$ lying within the boundary of the soft state referenced by the existing augmented state $\mathbf{x}_i$ assuming that their action vectors are approximately equal. That is, a high probability computed by (17) effectively indicates that executing the (local) policy embedded within $\mathbf{x}_*$ will very likely lead to a similar fitness value associated with the past experience $\mathbf{x}_i$. If, however, the underlying state vectors associated with $\mathbf{x}_i$ and $\mathbf{x}_*$ are sufficiently close within the reach of an action, then a high probability by (17) also implies their spatial proximity in the state space.

Policy interpolation in this manner indeed is reminiscent of that in Fig. 3 where the control policy for a new point is determined largely by the policy referenced by its neighboring particles. Note that (17), similar to (15), implicitly models a conditional probability density over the learning history due to the fact that the hyperparameters, shared by each kernel term $k(\mathbf{x}_i, \mathbf{x}_*)$, are optimized via the ARD process over the functional data set $\Omega_m$, which represents an up-to-date learning history. Consequently, the density estimation using (17) also depends on the past experience through the optimized hyperparameters. As we will see shortly, the experience association and the soft state transition are the foundation to further compressing the representation required for policy learning. However, the way the experience association is implemented in this paper, for simplicity, will adopt a deterministic scheme mentioned earlier that quantifies the association between two experience particles through a threshold parameter τ. Further analysis of the probabilistic model and its empirical study are left to the future versions of GRL implementations.

## V. CONCEPT-DRIVEN LEARNING ARCHITECTURE

The earlier sections generalize the RL framework by introducing the notion of the parametric-action model and its derived functional construct, action operator. This extended action formulation facilitates pattern discovery from within the augmented state space and thereby, serves as an instrument for identifying similar localized control policies through the process of experience association. As the kernel function in the GPR adapts to the newly acquired evidence during the course of policy search and the periodic ARD procedure, the reinforcement field also gets progressively evolved as a time-varying vector field in RKHS, all of which are influenced by the ongoing TD learning within the SARSA engine at the base layer.

From the angle of the reinforcement field, particles in $\Omega$, representing past learning experience, are encapsulated within the kernel functions (see (12)), each of which serves as an element that generalizes across the field for value predictions and policy inference for new particles. Since the reinforcement field is established based on the functional data view of the working memory that inherits properties of the RKHS, a decision-specific conceptual model (as in how past decisions are related to one another) can be developed through applying functional data analysis on the experience particles, each of which essentially holds a snapshot of the memory comprising the state and the action, with values manifested as their interactive random processes unfolded, along with their functional response as the fitness degree of their binding.

To discover coherent patterns hidden in such functional data, this paper proposes a simple union of Gaussian process regression (GPR) and spectral clustering as a *functional clustering* algorithm. With the operation of experience association, the concept-driven abstraction can be shown to reflect the correlation between decisions made across the state space and serve as a mechanism to reduce the decision points per state. This is useful particularly in complex domains with a larger action space that has non-stationary outcomes (e.g., task assignments in a large Grid network [20], [21], [23] with varying resource capacity). Moreover, each abstraction as a policy-embedding conceptual unit effectively serves as a simple cognitive model that can be shown to reveal the correlation between the state features and the action parameters. Such functional structure that goes across the state and the action space enables the agent to derive a high-level control policy by associating the past memory within each conceptual unit (rather than over the raw action set), giving rise to a prototypical model for the agent to "understand" the policy being followed.

### A. Clustering Functional Data via Gaussian Processes

GPR identifies the functional structure that relates correlated particles through their correlated fitness values. With this property, it opens the possibility to aggregate related past instances of memory, using the very same kernel as a similarity measure, to form clusters that consist of similar decision contexts (leading up to similar fitness values). For instance, the 3 query points marked by crosses in Fig. 3 along with their nearby particles in dots can potentially be grouped into 3 distinct clusters, each of which represents a coherent decision concept. Particles in these 3 clusters reference approximately identical fitness values due to the symmetry with respect to the final goal in this scenario (i.e., reaching and grabbing the flag). The point A, B, and C are approximately at equal distances to the flag and will remain so after the optimal actions are applied (i.e., action 2, 5 and 10 respectively).

Given a set of experience particles (i.e., $\Omega : \{(\mathbf{x}_1, q_1), ..., (\mathbf{x}_m, q_m)\}$), the goal is thus to identify such symmetry property within their corresponding localized control policies. Each cluster serves as a conceptual unit representing a guideline in choosing appropriate actions for all the states with sufficient correlations. In particular, the policy for a new query state can be inferred by matching particles within the same cluster via the experience association operation.

Recall from Section IV that a set of hypothetical states can be formulated by fixing the action while varying the state vector (and symmetrically, by fixing the state while varying the action vector). Here, to find the best match in the cluster, one seeks to find the most correlated particle(s) by comparing the state vectors between the query point and the target particles in memory ($\Omega$) while keeping the action vector fixed. Essentially, the agent asks this question: suppose I decide to take this action at my current state, are there any other similar historical contexts (or instances of memory) where taking this action leads to very similar results? By this token, the most correlated particle(s) for the query point A in Fig 3, for instance, are those nearby particles referencing action 2 as the best-known control policy. In other words, by taking the 2-o'clock action at A, the resulting particle (now referencing a hypothetical state) will become sufficiently correlated to the existing particles and thereby, the agent will be able to conclude that heading in the 2-o'clock direction must also be the best choice at point A.

One advantage of clustering correlated local policies as mentioned is to reduce the number of decision points per state when the set of primitive actions is large (e.g., dealing with a huge resource pool in the task-assignment domain). More importantly, forming conceptual units over policies also helps to identify the relationships between groups of states and actions that jointly lead to similar results (in terms of fitness measures). This allows for the agent to develop a cognitive model over the past decisions on top of the ongoing decision process itself. As an example, the section of empirical study will present a task assignment domain in which we will see that formulating decision concepts is helpful in automatically determining the optimal match-matching criteria between tasks and computational resources that underly most resource allocation and scheduling problems [20].

Moreover, since the clusters can be used as decision choices for deriving the control policy (by virtue of experience association), such policy abstraction can be made action-specific and considered as a type of action abstraction. Each abstract action is context-dependent due to the dependency over the underlying states in which actions were applied. Because the desired cluster structure needs to reflect both the correlation in the input data and their corresponding functional responses, such action abstraction can be established through functional clustering.

Functional clustering techniques have been used extensively, for instance, in identifying functionally similar genes from microarray data [32]. Methods of clustering functional data typically aims to "factor" the complex data distribution into a mixture of simpler models such as mixtures of regression models [33], infinite mixture model based on Dirichlet Processes [32], etc. By the rationale of data reuse and reduction of time and space complexity, this paper simply repurposes the covariance matrix computed from the GPR by reinterpreting it as a graph, which is then served an input for the spectral clustering mechanism to form functional clusters. This method is referred to as GP-Spectral Clustering (GPSC) in the later discussion. With GPSC, the goal is to cluster different areas in the state space that share similar (localized) control policies represented through the state-action pairs with correlated fitness values, in which the correlation is measured by the kernel function that powers the GPR.

*B. The Role of Spectral Clustering*

Spectral clustering has emerged recently as a popular method especially for discovering the patterns that are not easily captured by mixture models (e.g., mixture of Gaussians) or k-means due to unknown yet complex data distribution that goes beyond Voronoi sets (in which clusters shaped like "bubbles"). The main idea of spectral clustering is to identify cluster structures using eigenvectors of the similarity (or affinity) matrix derived from the data. Thus, all that is required to define clusters is to have a pair-wise similarity measure applicable for the data set. This characteristic also carries over to the case where feature vectors of the data are of varying dimensions, in which case, the mixture models or K-means will fail to apply.

Fortunately, the kernel used earlier as a correlation hypothesis over the augmented states can be reused again as the similarity measure. The covariance matrix $K$ from GPR effectively relates the input augmented states to their corresponding fitness value estimates. To formulate clusters of correlated local policies, it suffices to identity the cluster structure by reinterpreting the covariance matrix $K$ as representing a fully connected similarity graph with weights evaluated by the same kernel. To induce a sparser graph, we can again employ a threshold $\tau'$ by which the graph only retains the edges with similarity degrees greater than this threshold.

Note that the GPR's associated covariance matrix is a suitable choice for graph clustering purposes also due to the property of being symmetric and positive definite. Since highly correlated augmented states effectively represent similar decision contexts that led to correlated fitness values, the goal is therefore to identify a set of graph partitions such that the augmented states within the same partition have relatively higher probability to remain correlated to each other upon their soft state transitions (i.e., transitions by jumping from one correlated state to another covered in Section IV.B). This also implies that any new particle that is highly correlated to an existing partition (either in a complete-linkage or average-linkage sense) will be merged into the same partition with a high probability.

Formally speaking, suppose that the entries in the covariance matrix $K$ represents the weights in a fully-connected similarity graph denoted by $G = (\Omega, E)$. The vertex set represents the experience particles (i.e. $\Omega: \{(\mathbf{x}_1, q_1), (\mathbf{x}_2, q_2), \dots, (\mathbf{x}_k, q_k)\}$) and each edge in $E$ represents a degree of correlation between any two particles. Here we adopt a notational overlap with the functional data set used earlier for simplicity. Since the correlation between any two augmented states, i.e., $k(\mathbf{x}_i, \mathbf{x}_j)$, is calculated through maximizing the marginal likelihood (see (8)), a process that underlies the GPR, it also reflects the correlation in these particles' associated fitness values. Consequently, the similarity graph $G$ implicitly encodes a set of graph partitions with similar local policies (as entailed by functionally correlated particles).

Specially, imagine that we are given a new experience particle, posed as a query for its cluster membership. If at least a subset of the experience particles in the same partition is found to be more correlated to the new particle than those from the other partitions, then its cluster affiliation can be determined. As this dynamic process goes on, each partition will eventually contain sets of correlated particles with similar fitness values; equivalently, these particles have lesser tendency to escape from the same group according to their local policies (via the notion of soft state transitions). Consequently, the control policy represented through the experience particles in the same graph partition corresponds to a coherent decision concept.

For reasons above, the random walk graph Laplacian [17], denoted by $L_{rw}$ is an ideal candidate for grouping similar experience particles in that $L_{rw}$ effectively encodes a Markov transition matrix. To see this, notice first that one can define $L_{rw}$ in terms of the covariance matrix $K$ as a normalized graph

Laplacian: $L_{rw} = D^{-1}(D-K) = I - D^{-1}K$, where the *degree matrix* $D$ (i.e. $D = diag(d_1, d_2, ..., d_m)$) contains diagonal elements defined as follows:

$$d_i = \sum_{j=1}^{m} K_{ij} = \sum_{j=1}^{m} k(\mathbf{x}_i, \mathbf{x}_j). \quad (18)$$

Each diagonal entry in $D$ can be interpreted as the total weights (or correlation degrees) from the *ith* experience particle to all the other particles in $\Omega$. On the other hand, a first-order Markov transition matrix can be defined as $P = D^{-1}K$ such that

$$P_{ij} = \frac{K_{ij}}{\sum_{j=1}^{m} K_{ij}} = \frac{k(\mathbf{x}_i, \mathbf{x}_j)}{d_i}. \quad (19)$$

Since each row in $P$ sums to 1, $P_{ij}$ can be interpreted as the probability of transitioning from between states in $\mathbf{x}_i$ and $\mathbf{x}_j$ by executing their corresponding actions if the states are mutually reachable by the actions. For the cases where $\mathbf{x}_i$ and $\mathbf{x}_j$ are spatially further apart beyond the reach of an action, $P_{ij}$ can still be interpreted as a normalized degree of correlation. It is worth noting that, for a localized kernel, the formulation by (19) is analogous to the Nadaraya-Watson estimator given earlier in Section IV.B (see also (17)).

By (19), the relation between $L_{rw}$ and $P$, i.e., $L_{rw} = I - P$ implies that, if $(\lambda, \mathbf{v})$ represents the eigenvalue and associated eigenvector for $P$, then $(1-\lambda, \mathbf{v})$ corresponds to the 2-tuple for $L_{rw}$. As a result, the largest eigenvectors of $P$ or equivalently the smallest eigenvectors of $L_{rw}$ can be used to express the same cluster properties. The idea behind using eigenvectors as cluster indicators comes from applying *spectral relaxation* [16] – [18] to the graph-partitioning problem with the associated graph-based objective functions. In particular, the normalized cut [18] is among the most used measures of separation between clusters in graph and has been shown to have an equivalent relation expressed in terms of the Markov transition matrix.

As a simple illustration, first one defines the normalized cut in terms of the total links (or weights) between two graph partitions. Specifically, suppose that we wish to group particles in $\Omega$ into $p$ partitions; namely, $\Omega: \{\Omega^{(1)}, \Omega^{(2)}, ..., \Omega^{(p)}\}$, where $p$ in general is chosen to be much smaller than the sample size of the training set $m$ (i.e., $p < m$). The total weights between $\Omega^{(i)}$ and $\Omega^{(j)}$ can thus be defined as follows:

$$W(\Omega^{(i)}, \Omega^{(j)}) = \frac{1}{2} \sum_{\mathbf{x} \in \Omega^{(i)}, \mathbf{x}' \in \Omega^{(j)}} k(\mathbf{x}, \mathbf{x}'), \quad (20)$$

where the factor ½ is introduced such that, each link is counted only once. Then, the *p*-way normalized cut can be defined as:

$$NCut(\Omega^{(1)}, \Omega^{(2)}, ..., \Omega^{(p)}) = \sum_{i=1}^{p} \frac{W(\Omega^{(i)}, \Omega \setminus \Omega^{(i)})}{W(\Omega^{(i)}, \Omega)}, \quad (21)$$

where $\Omega \setminus \Omega^{(i)}$ denotes the complement of $\Omega$ at the *i-th* partition, and the denominator $W(\Omega^{(i)}, \Omega)$ can be thought of as the sum over all the internal and external links; namely,

$$W(\Omega^{(i)}, \Omega) = W(\Omega^{(i)}, \Omega^{(i)}) + W(\Omega^{(i)}, \Omega \setminus \Omega^{(i)}).$$

Thus, for obtaining p-way partitioning of $\Omega$ using the normalized cut formulation, one seeks to solve the following optimization problem:

$$\text{minimize } \frac{1}{p} NCut(\Omega^{(1)}, \Omega^{(2)}, ..., \Omega^{(p)}). \quad (22)$$

The optimization objective in (22) effectively encourages a balanced graph partition (due to the denominators within the summation in the normalized cut in (21)) that in the meantime, also minimizes the total escaped links across the partitions (due to the numerators in (21)). However, finding the solution to (22) is NP-hard [34], thus off-the-shelf spectral clustering algorithms often resort to spectral relaxation. In particular, the optimization problem with a normalized cut criterion in (22) can be simplified to the following trace maximization problem [18]:

$$\text{maximize } \frac{1}{p} tr(Z^T K Z), \quad (23)$$

where $Z = X(X^T D X)^{-1/2}$ and $X$ is an $m \times p$ matrix with each column as a partition indicator (i.e., a binary vector with entries 1 representing the partition membership and 0 otherwise). Note that $K$ is the original covariance matrix from which the weights between any two partitions in (20) are calculated and $D$ is the degree matrix defined earlier. With the definition of $Z$, a simple matrix multiplication shows $Z^T D Z = I$ is an implicit constraint for (23), where $I$ is an $p \times p$ identity matrix. For the solution to be tractable, (23) can be approximated by using the top $p$ eigenvectors of $L = D^{-1/2} K D^{-1/2}$ with the constraint $Z^T D Z = I$ relaxed such that, $Z$ as a function of $X$, is allowed to assume continuous values (instead of 0s and 1s). Expressing the solution in terms of the eigensystem $L\mathbf{v} = \lambda \mathbf{v}$ and multiplying both sides by $D^{-1/2}$ gives $D^{-1} K D^{-1/2} \mathbf{v} = \lambda D^{-1/2} \mathbf{v}$, which can be written to be:

$$P(D^{-1/2} \mathbf{v}) = \lambda (D^{-1/2} \mathbf{v}).$$

Therefore, the top $p$ eigenvectors for the system $L\mathbf{v} = \lambda \mathbf{v}$ also corresponds to the top $p$ eigenvectors for the Markov transition matrix $P$ only they are scaled by a factor of $D^{-1/2}$. Lastly, given that $L_{rw} = I - P$, the smallest $p$ eigenvectors are used instead to compute the a discrete partitioning of $\Omega$.

With the partition set $\Omega: \{\Omega^{(1)}, \Omega^{(2)}, ..., \Omega^{(p)}\}$ derived from the spectrum of the random walk Laplacian $L_{rw}$, we can then interpret each subset $\Omega^{(i)}$, comprising a set of correlated experience particles, as an abstract action $A^{(i)}$, given that these instances of memory share relatively similar decision contexts, representing local policies with correlated fitness values. This paper uses the Shi-Malik flavor of the spectral clustering algorithm [34] to obtain these graph partitions with the assumption that the number of the abstract actions is pre-specified as a hyperparameter. Incidentally, we know that, each abstract action defined as such immediately represents contextually correlated particles with similar fitness values, also due to the property of the covariance matrix $K$ in GPR, which is used to define the graph Laplacian $L_{rw}$. Graph partitioning thus serves as a basis for defining the corresponding abstract actions denoted by $A^c: \{A^{(1)}, A^{(2)}, \dots A^{(p)}\}$.

Meanwhile, recall from Section IV.A where we have defined a hypothetical state $s_h^+$ effectively as a query point in the state space with optimal actions yet to be determined by running experience association from within the working memory. In order to equip the agent the capacity to infer a local policy for a new hypothetical state using the nearby particles (e.g., finding optimal actions for query points A, B and C in Fig. 3), we will again represent the policy function $\pi$ through the fitness function $Q^+(\cdot)$, the same way as the case of parametric actions, where $\pi$ takes the form of a functional $\pi[Q^+]$ with the probability of an action increases over $Q^+$ (see (10)). Since the value of $Q^+$ is driven by the experience particles in $\Omega$ (see (12) and (16)), the policy also depends on the progressively updated memory via particle reinforcement (Section IV).

With the control policy represented through $Q^+(\cdot)$, each abstract action $A^{(i)}$, being effectively a graph partition, now represents a high-level strategy encompassing multiple localized, yet related, control policies $\{(\mathbf{x}_s, \mathbf{x}_a)_j^{(i)}, q_j^{(i)}\}$, in which the constituent state and action vectors $(\mathbf{x}_s, \mathbf{x}_a)_j^{(i)}$ and their fitness measure $q_j^{(i)}$ are taken from the $j$-th particle in the $i$-th partition, i.e., $\omega_j^{(i)} \in \Omega^{(i)}$. Correlated particles in memory with respect to the hypothetical state in question can therefore be resolved into a set of candidate primitive actions. Specifically, this can be achieved by extracting the action vectors $\{(\mathbf{x}_a)_j^{(i)}\}$ from these correlated particles and applying the parametric-to-primitive action map, $f_A: A^+ \to A$, a task-dependent function with constraints on action parameters, to obtain their corresponding primitive actions $\{a^{(k)}\} \subseteq A$, where we have used a different index $k$ for the primitive action to distinguish from the indices for particles and graph partitions.

Intuitively, if a query state lies in the proximity of those states referenced by a particle set, then applying the actions inferred from these particles will very likely lead to similar results as indicated by their fitness values. The example in Fig. 3 provides a simple way to visualize the policy inference as such, where the agent at query points A, B and C would infer the appropriate clock directions to navigate according to their nearby particles. We can also view the inference from the perspective of the soft state transition, for which an example can be observed from the bottom half of Fig. 5, where each soft state area encloses highly correlated particles that lead to the successor soft state, which is represented by yet another set of correlated particles.

Using graph partitions, we can in principle establish an $n$-tier hierarchical inference system such as a 3-tier hierarchy where we start with the set of abstract actions $\{A^{(i)}\} \subseteq A^c$ (tier 1), each of which references a set of soft states (tier 2), which themselves reference a set of correlated particles (tier 3). Fig. 6 shows a formation of two abstract actions established by running a selected graph partitioning algorithm on experience particles in the working memory. At the bottom of Fig. 6, each abstract action is represented by a group of correlated particles within the same graph partition, which may contain finer subgroups of particles forming stronger bonds.

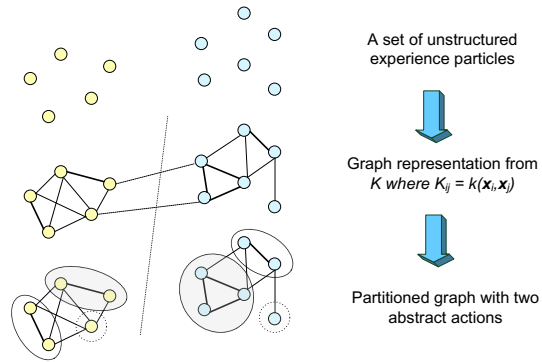

Fig. 6. Abstract action hierarchy in CDLA. Experience particles in the working memory form a functional data set as the byproduct of the GPR characterized by a mean and covariance function; they can also be interpreted as a graph with an associated similarity matrix $K$ implicitly defined through the covariance matrix. Running spectral clustering on $K$ therefore produces a set of (approximate) graph partitions, each of which references a set of correlated particles with correlated state-action bindings and fitness values.

Note that in general, the correlated particles need not reference spatially continuous state/action vectors. This is depicted by the dotted ellipses at the bottom of Fig. 6, in which each ellipse (in the same partition) references a set of spatially connected particles (in terms of their referenced states) but nonetheless is apart from those particles in the other ellipses. For instance, the 3 discontinuous regions in Fig. 3 can potentially be grouped into one partition (depending on the kernel of choice and its hyperparameters) due to their symmetry resulted from their correlated fitness values, being equally apart from the goal state. In general, different state-action bindings (representing localized policies) may be grouped into the same abstract action due to their representative particles having similar fitness values. As mentioned in Section IV.B, particles are said to form weak bonds if they are correlated, therefore coupled, by experience association but are not mutually reachable; groups of these particles as soft states can make "weak-sense" soft state transitions through memory association. On the other hand, particles that are both coupled (by association) and mutually reachable (via some actions) are said to form strong bonds. In Fig. 3, particles near the same query point may form strong bonds due to their mutual reachability with an action, whereas particles at different query points (say, A and B) are correlated but not reachable, therefore can only form weak bonds.

An abstract action derived directly from a single application of spectral clustering, without further treatments, may contain both groups of weakly bound and strongly bound particles under the measure of the kernel function. Ultimately, the RL agent needs a primitive action to move within the state space and therefore, an abstract action as a choice must be reducible to a concrete, primitive action. To that aim, one can make use of the clustering property and the operation of experience association to devise a systematic matching process, simply referred to as *action resolution* (from the abstract to the primitive).

Specifically, the agent at the current state makes a query into an abstract action by performing experience association. The query is formulated by forming a hypothesis in terms of a hypothetical state $s_h^+$, which, just like a regular augmented state, also consists of a state vector and an action vector; the state vector is the agent's current state, and the action vector is inferred from a matched particle as a member of the abstract action. If such a hypothetical state is sufficiently correlated to at least some particles referenced by a target abstract action $A^{(i)} \in A^c$, then $A^{(i)}$ effectively specifies, for the query state, an action-resolution strategy that resolves into a set of primitive actions from the matched particles. That is, the matched particle in $A^{(i)}$ would reference a state vector and an action vector that, when combined (denoted $s^+$), form a bond with the hypothetical state as a particle such that $k(s_h^+, s^+) \geq \tau$. Note that we have used a linguistic shortcut to refer to a hypothetical state as a "particle" even though it is still missing a fitness measure, which is to be inferred by the GPR layer.

From the decision context perspective, consider a state $s \in S$ in an experience association process with target particles referenced by an abstract action $A^{(i)} \in A^c$. If $A^{(i)}$ indeed references at least one correlated particle with respect to $s$, then $A^{(i)}$ is said to be *in context* with $s$; otherwise, $A^{(i)}$ is said to be *out of context*. If the chosen abstract action is in context with the agent's current state and in the meantime also resolves into a primitive action leading up to a high fitness value, then such abstract action is given a relatively high preference according to the policy representation as an increasing functional of $Q^+$, i.e., $\pi[Q^+]$. We will generalize the policy function (10) shortly in Section C to take into account abstract actions.

One caveat arises when the chosen abstract action $A^{(i)}$ is out of context, not referencing any particles sufficiently correlated with the agent's current state: by applying experience association with candidate particles in $A^{(i)}$ results in hypothetical states $\{s_h^+\}$ such that they all fall short of the threshold-based criterion, i.e., $k(s_h^+, s^+) < \tau$, for all $s^+$ induced from particles in $A^{(i)}$. Since no points of reference exist in such cases, a randomized action-resolution strategy is adopted as a backoff for these query states. The implication of random policy is to be discussed shortly.

Action resolution can be a complex model and does not need to be deterministic. Multiple matched candidates for a query state can occur, which, when all considered, can lead to multiple candidate trajectories. For simplicity, however, a greedy policy is used in this paper to obtain a single primitive action via experience association. The greedy policy will always favor the primitive action inferred from the most correlated particle, with which the parametric-to-primitive mapping $f_A: A^+ \to A$ is applied to its action vector, while factoring in constraints, to solve for the corresponding primitive action of interest.

Since the main objective of experience association is to infer the best action at a query state by sorting through correlated past memory instances (within the same graph partition), we can assume that the same action, as entailed by the matched particle, will be replicated at the query state. That is, for the abstract-to-primitive action resolution, the experience association is effectively determined by the correlation of the state (i.e., the context) between the query and the past memory instance, while holding the assumption that the action vectors will remain approximately identical, i.e., the uncertainty in the action parameters is negligible between the hypothetical state (the query) and the matched particle (the past memory).

In the use case where the hypothetical state (as a particle) is required to form a strong bond with the candidate particle, we also need to consider an additional constraint ensuring that the state between the query and the target memory instance are mutually reachable by the action inferred from the memory. In the case where the experience association is applied with respect to a large population of particles (e.g., a sizable abstract action), a sample-based strategy can be used to prevent from an exhaust search to reduce the computation cost. In particular, the particle reinforcement algorithm introduced earlier in Section IV.A based on the local search, a search within the same state space partition, can be reapplied in action resolution to effectively reduce the search space.

### C. Fitness Value Estimate for Abstract Actions

With the both experience association and the concept-driven abstract actions established, each augmented state finally has an intermediate representation from which we can define its fitness value. Recall from section II.E that a fitness value is essentially a utility estimate obtained from the base layer (running, say, the SARSA update rule) that gets propagated to the GPR layer as a training signal in response to the bonding between a state and an action (in its parametric form). Under the model of the action operator, uncertainty underlying the action parameters are resolved only when the operator acts on the state (see (11) for an example), and the resulting action vector $\mathbf{x}_a$ combined with the state vector $\mathbf{x}_s$ in turn forms an augmented state $(\mathbf{x}_s, \mathbf{x}_a) \in S \times A^+$. The observed reward $r$ is used to update the utility estimate for this state-action pair at the base control learner, which is not aware of the underlying action parameters. The only additional step beyond the usual utility update is to relay this utility estimate as the target fitness value for the corresponding augmented state, which altogether forms a new experience particle represented by the 3-tuple $(\mathbf{x}_s, \mathbf{x}_a, q)$ and subsequently include it into the agent's working memory as a functional data set $\Omega$.

A fitness value can be interpreted as an approximate utility estimate that assimilates the uncertainty during the update history of the associated state-action bonding (i.e., the iterative Q-value estimate at the base level). Since the augmented state carries extra degrees of freedom from the action parameters, using a utility as a measure for a given augmented state effectively performs a "smoothing" over the stochastic effects inherent in the action behavior. Similarly, the fitness value estimate for a pair of state and abstract action is effectively a further averaging over the sequence of fitness values $(q)_{j=1...m}$ inferred from the action-resolution history of size $m$ that unfolds as sequential pairs of state and parametric-action $(\mathbf{x}_s, \mathbf{x}_a)_{j=1...m}$, which in parallel are reduced to the state-and-primitive-action pairs $(\mathbf{x}_s, a^{(i)})_{j=1...m}$ via $f_A: A^+ \to A$, where $\mathbf{x}_a \in A^+$ and $a^{(i)} \in A$ as before.

Although using the averaged utility as the fitness value estimate can only serve as an approximation (for which more accurate estimate from within the augmented state space is an open research question), the role of the kernel in the GPR prediction allows for a reasonable generalization of value prediction to occur across the augmented states. The predicted fitness value will remain accurate so long as some experience particles exist in the neighboring space of the query point. This can be seen from (16) where the mean prediction of the GP is expressed through the kernels centered at all the other augmented states; that is, fitness value prediction for any new augmented state is largely determined by those particles with high correlations. Effectively, the prediction generalizes into the unknown parts of the state space through the kernel as a correlation hypothesis. By (7), we also get that the more neighboring particles exist in the query point, the less the variance of the prediction will be.

The initial fitness value estimate for a given augmented state is the reward discounted by the learning rate (e.g., $\alpha R$ in (2)), assuming that all Q-values are initialized to 0. The same 2-tier architectural approach for the fitness value estimation in the reinforcement field with the base layer carrying out a standard TD learning, such as SARSA, can be adapted to estimating the fitness value under the abstract-action model. However, to obtain a set of graph partitions (as the abstract action set $A^c$), at least a few experience particles must be gathered as seeds. This can be achieved by following a random policy for some time steps (i.e., pure exploration) to seed an initial set of experience particles, from which we can then jump start the very first action abstraction model. The only way for the "infant agent" to formulate an initial cognitive model of the world is to perform certain trials and errors to gather hands-on experience. Alternatively, it is also possible to establish the initial graph partition with a prior cluster assumption based on the similarity between augmented states. Note that using a random policy in such cases (i.e., at initial query states) corresponds to the action resolution from out-of-context abstract actions that carry no relevant contexts (i.e., states) for the agent to perform experience association since the agent effectively has no points of reference.

As mentioned earlier, a greedy policy can be used to implement the action resolution (from the abstract to the primitive) by systematically running the kernel-based criterion on the past memory instances within relevant decision contexts as represented by the abstract actions. In the abstract action model, the decision-making process unfolds in a hierarchical fashion with the agent learning an abstract-level control policy over $\{A^{(i)}\} \subseteq A^c$ while the selected abstract action $A^{(i)}$ is reduced to a primitive action $a^{(i)} \in A$ through experience association. Following the policy over $A^c$, the agent subsequently executes the primitive action resolved from the selected abstract action, which, according to greedy action resolution, always chooses the one that leads to the highest-valued augmented state. The agent then receives a reward and updates the utility estimate with respect to the given pairing of state and abstract action, say, $(s, A^{(i)})$, by following the update rule of the TD algorithm at the base level. In the case of SARSA, this amounts to a generalization over (2) by redefining the two-step horizon TD update rule respect to $\{A^{(i)}\}$:

$$Q(s, A^{(i)}) \leftarrow Q(s, A^{(i)}) + \\ \alpha[R(s, A^{(i)}) + \gamma Q(s', A^{(i)}_{next}) - Q(s, A^{(i)})] \quad (24)$$

The utility estimate for the $(s, A^{(i)})$-pair, denoted as $q_i$, is in turn used as a fitness value for the corresponding augmented state $s^+ = (\mathbf{x}_s, \mathbf{x}_a^{(i)})$. This produces one complete experience particle: $\omega_i = (s_i^+, q_i)$, where $q_i$ is an estimate for $Q(s, A^{(i)})$.

The moment an abstract action $A^{(i)}$ is chosen by the agent in context with its query point $s$, where a hypothesis state $s_h^+$ is made, that decision hypothesis $s_h^+$ is, by definition, guaranteed to be correlated to at least a subset of existing particles associated with $A^{(i)}$ through experience association. The value $Q(s, A^{(i)})$ is subsequently passed on to the GRP layer to serve as a target fitness value for the augmented state $(\mathbf{x}_s, \mathbf{x}_a^{(i)})$, where $\mathbf{x}_a^{(i)}$ is the parametric form of the primitive action $a^{(i)}$ as a reduction from an abstract concept $A^{(i)}$ under experience association. We can also justify such value approximation by the fact that correlated inputs (i.e., the set of augmented states $\{s^+\}$ or $\{(\mathbf{x}_s, \mathbf{x}_a)\}$) for the GPR, as a value function approximator, always get mapped to correlated functional outputs $\{Q^+(s^+)\}$ as dictated by the GPR's kernel function as a measure of covariance – and it is also based on the same kernel function that the graph partitions are derived and interpreted as abstract actions.

On the other hand, if the chosen abstraction $A^{(i)}$ is out of context with the agent, the value $Q(s, A^{(i)})$ may not be used directly as an estimate for $Q^+(s^+)$ in that $Q(s, A^{(i)})$ only reflects a random choice rather than the utility estimated from correlated particles as a decision concept. One could attempt to estimate $Q^+(s^+)$ in such out-of-context scenarios by performing experience association against all the known particles in $\Omega$ to find the most correlated instances of memory and their associated fitness values; however, in the extreme cases, they may not exist at all under the kernel-based threshold. Again, maintaining a set of state partitions can effectively reduce the experience association at the global scale into a local search within a relevant state partition. Details for the fitness value estimate under such degenerative, out-of-context cases are covered in the next section.

Note that the action resolution from the abstract to the primitive action will gradually tend to a deterministic process as the underlying graph partitions tend to a stable set, which promotes the convergence of $Q(s, A^{(i)})$. This can effectively be controlled by a gradual decrease of the rate of cluster reformulation (e.g., from once per $T$ state transitions to once per $T + f(T)$ transitions) as the agent gathers experience particles and runs particle reinforcement, periodically updating the working memory. As alluded earlier, the control policy $\pi$ under

| **Algorithm 3: G-SARSA(** $k, p, T, \Omega_0, \tau$ **):** |
|---|
| **Input:** |
|     $k$: Kernel function $k(\cdot,\cdot)$ with hyperparameters $\boldsymbol{\theta}_0$ <br>     $p$: Number of abstract actions <br>     $T$: Cycle for model reformulation (used by GPSC) <br>     $\Omega_0$: Initial experience particles <br>     $\tau$: Threshold of correlation (lower bound) |

1. **Bootstrap** an initial abstract-action model:
   $A_0^c: \{A_0^{(i)} | i = 1 \ldots p\}$ using an initial sample set $\Omega_0$ via **GPSC**
   // see Section V for details on GPSC
2. **Initialize** $Q(s, A^{(i)})$ arbitrarily
   2.1 Initialize $Q(s, a^{(i)})$ arbitrarily
   // In general, $Q(s, a^{(i)})$ can be different from $Q(s, A^{(i)})$
   // In this basic G-SARSA, both Q-values are always the same,
   // since $A^{(i)}$ is always reduced to a single primitive action $a^{(i)}$.

**Repeat** (for each episode)
3. **Initialize** $s$.      //starting state
4. **Choose** $A^{(i)} \in A^c$ from $s$ using $\pi(s, A^{(i)}) = \pi[Q^+]$,
   where $\pi$ is the softmax policy given by (25), and apply **experience association** with respect to $(s, A^{(i)})$, which reduces $A^{(i)}$ to $a^{(i)} \in A$ (with parametric form $\mathbf{x}_a^{(i)} \in A^+$) under an action-resolution policy (e.g., greedy); the corresponding augmented state $s^+ = (\mathbf{x}_s, \mathbf{x}_a^{(i)})$

**Repeat** (for each step in the episode)
5. **Run ARD**, and **reformulate** abstract action set $A^c$ using **GPSC** once in every $T$ state transitions
   5.1 Optionally update T,
      $T \leftarrow T + f(T)$, where $f$ maps $T$ to some number as an increment for $T$.
6. **Take action** $a^{(i)}$ observe $r$ and $s'$
7. **Choose** $A^{(j)} \in A^c$ from $s'$ using $\pi[Q^+]$, where $A^{(j)}$ reduces to $a^{(j)}$ with parametric form $\mathbf{x}_a^{(j)}$ through applying **experience association** on $(s', A^{(j)})$
8. **Update** $Q(s, A^{(i)})$ by applying the update rule (24) with the observed reward $r$ from Step 6 and $A^{(j)}$ from Step 7

   8.1 $Q(s, a^{(i)}) \leftarrow Q(s, A^{(i)})$ // $a^{(i)}$ is reduced from $A^{(i)}$
   8.2 **Save** the new Q-value estimate for later use
      $q_i \leftarrow Q(s, a^{(i)})$ // fitness estimate for $(\mathbf{x}_s, \mathbf{x}_a^{(i)})$
   8.3 **Save** the temporal difference, i.e.,
      $TD(s, a^{(i)}) = [R(s, a^{(i)}) + \gamma Q(s', a^{(j)})] - Q(s, a^{(i)})$
      $t_i \leftarrow TD(s, a^{(i)})$
9. **Propagate** the $Q$-value obtained from Step 8 to the GPR layer as the fitness value estimate for $(\mathbf{x}_s, \mathbf{x}_a^{(i)})$ from Step 7.
10. $s \leftarrow s'; a^{(i)} \leftarrow a^{(j)}$ // $a^{(j)}$ resolved from $A^{(j)}$ at step 5
11. **Update** $\Omega$ by running **particle reinforcement** with **MemoryUpdate(**$\omega, t_i, k, \tau, \Omega$**)**, where
   $\omega = ((\mathbf{x}_s, \mathbf{x}_a^{(i)}), q_i)$ and $t_i$ is an estimate for $TD(\omega)$

**Until** $s$ is terminal or max number of episodes reached abstract actions $A^c$ is also driven by the experience particles the same way as the policy learned via the RF-SARSA for the parametric-action-based control policy since in both cases, the policy is represented through the fitness function $Q^+(\cdot)$, which is represented by the GPR with the covariance function $k(\cdot,\cdot)$. Using a softmax function as the sampling strategy for selecting abstract actions $\{A^{(i)}\}$ along with the (negative) fitness (i.e., $Q^+(\cdot)$) as the energy function, we get the policy function $\pi[Q^+]$ over $A^c$, analogous to (10):

$$\pi(s, A^{(i)}) = \frac{\exp[Q(s, A^{(i)})/\tau]}{\sum_{j \in \{1,\ldots,p\}} \exp[Q(s, A^{(j)})/\tau]}$$

$$= \frac{\exp[Q^+(s_i^+|s, A^{(i)})/\tau]}{\sum_{j=\{1,\ldots,p\}} \exp[Q^+(s_j^+|s, A^{(j)})/\tau]} \quad (25)$$

where $p$ is the number of abstract actions and $\tau$ denotes a temperature (not to be confused with the kernel-based threshold for experience association). The term $Q^+(s_i^+|s, A^{(i)})$ in (25) denotes the fitness value at the hypothetical state $s_i^+$ given the agent at the query state $s_i$ performing experience association with $A^{(i)}$. The value of $Q^+(s^+|s, A^{(i)})$ is in general estimated through the GPR layer based on the particles in the memory (i.e. $\Omega$). A stripped-down version of the generalized SARSA (G-SARSA) is illustrated in Algorithm 3. On the top level, the G-SARSA is similar to the standard SARSA in terms of the Q-value updates; however, we need to factor in the systematically reduction of an abstract action to a primitive action in the policy search. We will unpack additional details and cover edge cases for G-SARSA in the next section.

### D. More on the G-SARSA

The vanilla G-SARSA outlined in Algorithm 3 may not work as it is until we zoom in and handle a few edge cases. Notably, what happens when the agent is out of context with the existing abstract actions, and how do we choose actions and estimate their fitness values in such cases (Step 7, 8 and 9)? How do we assign the cluster membership to new particles, and relatedly, how do we keep the cluster index as consistent as possible across multiple runs of the GPSC reformulating the clusters (Step 5)?

In Step 7 of the vanilla G-SARSA, the utility estimate $Q(s, A^{(i)})$ for the abstract action $A^{(i)} \in A^c$ is carried over to the GPR layer to serve as the fitness value for the corresponding augmented state $(\mathbf{x}_s, \mathbf{x}_a^{(i)})$, which is consistent with the 2-tier architecture depicted earlier in Fig. 2. Such fitness value estimate can only be reasonable under the assumption that the chosen abstract action is in context with the agent, i.e., being consistent with the agent's current state by experience association. This is because each abstract action is periodically bootstrapped as one of the graph partitions comprising correlated particles and therefore can only be used to match a correlated hypothetical state, which is then used to temporarily augment the existing graph partition as a new member. From the value estimation standpoint, only when the new particle is in context will the resulting estimate be consistently correlated

4with those of the other existing experience particles.

By contrast, a random action-resolution strategy that the agent is forced to follow at the out-of-context states (i.e., when the experience association failed to match any particles in the chosen abstract action $A^{(i)}$) will result in utility estimates that only reflect random choices of actions without the correlation structure (the graph partitioning) afforded by the action abstraction. In such "degenerative" cases, the hypothetical state is formulated by merging the value of the queried state with an action vector resolved from a random primitive action, which amounts to a pure exploration strategy. Note that the random action resolution also can occur in situations where an abstract action does not reference any particles (i.e., an empty cluster), which automatically defaults to the out-of-context scenario.

Choosing a random (primitive) action, say, any $a^{(i)} \in A$, may seem unpreferable in the presence of other valid, non-empty abstract actions, it is nonetheless a necessary exploratory mechanism for the agent to uncover desirable actions in the unknown decision contexts. Since the experience particle generated from pure exploration as such cannot be associated with any abstract action as a concept, we (might) need a different way to estimate its fitness value rather than using the Q-value directly from (24). The most straightforward method is to use the prediction from the GPR as the fitness value, i.e., $Q^+(s_h^+) = Q^+(\mathbf{x}_s, \mathbf{x}_a^{(i)})$, which is essentially an extrapolation from all the existing particles in the working memory. However, doing so will prevent the agent from incorporating the observed reward $r$ in Step 6, which helps to re-estimate the Q-value associated with the random action $a^{(i)}$.

By examining the role of Step 6 for the value update in Step 8 using (24), it may seem that $Q(s, A^{(i)})$ is influenced by a reward from a randomized action, which is not directly associated with the abstract action $A^{(i)}$ itself. On the other hand, we can think of this reward as both factoring in, probabilistically, the event of a success experience association, which then reduces the abstract action to a sensible primitive action leading up to a "true" reward $r_T$, and the event of a failed experience association, which defaults to a random action and consequently, a "random" reward $r_R$. In other words, such out-of-context reward, in the long run, is effectively a weighted average of the true reward and the random reward:

$$\beta r_T + (1-\beta) r_R,$$

where the probability of successful experience association $\beta$ is a running statistic as the working memory continues to be updated and the clustering structure continues to evolve. Each abstract action in principle can have a different probability of success for experience association. It is possible for an abstract action to have a small $\beta$, making it harder to be selected but at the same time, running experience association with such abstraction is helpful given that it carries relevant decision contexts. In this version of GRL, however, we will not address the implication of different values of $\beta$, which will be studied elsewhere. For simplicity, we will use the reward as it is from Step 6 to perform the value update in Step 8 in out-of-context cases, assuming that each abstract action will eventually have a fair share of randomness as such and the fitness value estimate will be dominated by successful experience associations as the agent learns to make progressively better decision, with high probability, among the choices of abstract actions.

As a new particle is introduced into the memory, we need to assign it a temporary cluster membership, which holds fixed until the graph partitioning is restructured at every $T$ steps in an episode (Step 5). How the cluster membership can be determined for new data is an open question, for which the solution can range from learning eigenfunctions [41] in the case of spectral clustering, computing the distances to cluster centroids and choosing the one with the shortest distance, to simply training a classifier using existing data. In this paper, we will focus on the *k*-nearest neighbors approach (kNN) for its simplicity, and consistency with the graph partitioning. The prediction steps can be summarized as follows:

1) For each candidate abstract action $A^{(i)} \in A^c$, compute the average degree of correlation ($\rho^{(i)}$) – between the new particle ($\omega$) and the *k* most correlated particles associated with a target abstract action – using the (periodically tuned) kernel function from the GPR layer.

2) Check the threshold of correlation (e.g., using the threshold $\tau$ from experience association). If the most correlated abstract action has an average degree of correlation $\rho^{(i)} \geq \tau$, then merge $\omega$ into the target abstract action. If $\rho^{(i)} < \tau$, then assign $\omega$ to an empty abstract action (i.e., empty cluster) if it exists; if, however, all abstract actions have existing members, assign $\omega$ to the most correlated abstract action anyway (and wait for a potential reassignment during the cluster reformulation at Step 5).

Incidentally, the cluster assignment for a new particle (with its state and action vectors fixed) is very similar to running experience association on a set of target particles when the agent, at a new state, makes policy inference drawing upon past memory by forming hypothetical states (Section IV.A). The only difference is that, in this case, the hypothetical state has been fixed and no extraction of action vectors from target particles is involved.

The cluster membership assignment via kNN is only temporary because we will need to periodically re-formulate the cluster. As the continuous stream of new particles is incorporated into the agent's memory, the cluster structure will evolve over time. The temporal difference update at Step 8 will constantly shift the Q-value estimates, which, batch by batch are propagated to the GPR layer to serve as target values (labels) for the fitness value prediction. Meanwhile, as the periodic ARD process changes the hyperparameters of the kernel function, the value prediction for new particles will change accordingly. Working synergistically with the GPR layer, the process of particle reinforcement constantly updates the working memory by replacing existing particles with new ones, reflecting new local policies and new value estimates.

The big picture is that the agent maintains a set of memory instances, called particles, that are constantly replaced by new pieces of evidence and new estimates of fitness. For this reason, we need to consider reformulating the clusters periodically to reflect the new evidence, which then gives rise to a renewed set of abstract actions. However, under G-SARSA, the goal is for the RL agent to learn a high-level control policy over abstract actions; the hope is that each abstract action, when renewed, will stay as consistent as possible such that it continues to represent a relatively steady decision concept without rapid concept drifts. One way to minimize the concept drift is to maximize the overlap between the old and the new cluster (with the same index) upon cluster reformulation. Specifically, given $\{A_t^{(i)}\} \subseteq A_t^c$ and $\{A_{t+1}^{(i)}\} \subseteq A_{t+1}^c$, we wish to assign the cluster index $i$ at step $t+1$ such that $A_{t+1}^{(i)}$ is as similar as possible to $A_t^{(i)}$ for all $A_{t+1}^{(i)}$ in $A_{t+1}^c$. This is yet another open question, for which we may define a cluster similarity measure, using the kernel function, and assign the new cluster index such that the highest similarity can be achieved for the cluster before and after its reformulation. A simpler method we consider in this paper is to keep track of the cluster index for each member particle and use majority vote to determine the new cluster index. Under the assumption that the cluster structure does not change drastically by setting an appropriate reformulation period $T$, most of the differences before and after re-clustering (or graph re-partitioning) would stem from the introduction of new particles and a slight shift in the kernel hyperparameters $\boldsymbol{\theta}$, which comes from the shift in the Q-value estimate. Systemic studies of concept drifts and impacts of cluster similarity are left for future research.

## VI. EMPIRICAL STUDY

In this section, we will first illustrate the reinforcement field as the primary RL generalization mechanism aiming to cope with complex control scenarios involving actions with multivariate parameters and fluid behaviors. In particular, we will demonstrate RF-SARSA by reusing the 2D navigation domain depicted in Fig. 4 but with different configurations and layouts. We will present how control policies can be learned from the parametric-action model along with the reinforcement-field construct and further show how the (functional) field can be visualized when the action vector sits in a lower dimensional space such as 2D (and 3D). We will also progressively make the navigation environment more complex (by introducing obstacles) and experiment on different choices of kernel functions, through which we then observe how the reinforcement fields would vary accordingly.

Next, to motivate the CDLA framework as a second RL generalization mechanism – policy learning with actions that themselves represent domain-specific concepts – we will examine the task-assignment domain briefly mentioned earlier in Section II.C, where the abstract actions, as decision concepts, are relatively straightforward to interpret. Additionally, the task assignment domain is perhaps a typical example, where having multiple action parameters simplifies the RL problem formulation and where forming concepts across decision contexts can further assist in shaping intelligent behaviors through concept-driven policy learning. Specifically, we will demonstrate the use of G-SARSA under the CDLA framework in solving an instance of task-assignment problem where the resource profile of the servers changes over time.

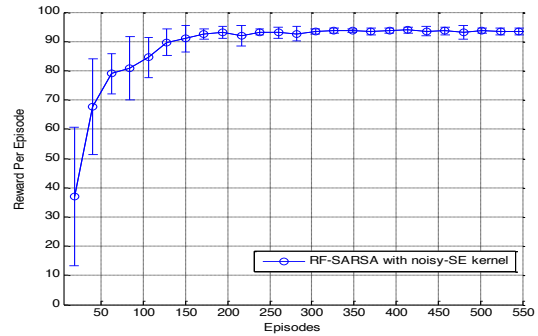

Fig. 7. Learning curve for RF-SARSA with noisy SE kernel.

### A. Reinforcement Field in a Navigation Domain

The first experiment investigates the simplest scenario where no obstacle exists in the state space such that the agent is free to roam around in all directions within the state space boundary. The state space is partitioned into 25 evenly spaced grids, each of which is configured to reference 3 positive and 3 negative particles respectively such that a maximum of 150 particles are maintained in memory. Since the goal is for the agent to learn the policy following a shortest path, a cost (-1) is imposed per step of travel whereas a reward (+100) is given upon reaching the goal area. Assuming that each side of the environment is of 5-unit length, the action parameter $\Delta r$ is set to have a target value of 1-unit length but varies between -0.8- to +1.2-unit length modulated by a Gaussian noise, applicable to all action choices. The angular parameter $\Delta \theta$ takes on 12 equal partitions around the circle (see also Fig. 1 and Fig. 3) resulting in 12 different clock directions for the actions, each of which has ±15 degrees of a tolerable error. Following a shortest path to the goal area will obtain a reward of approximately 94. Fig. 7 shows the learning curve from RF-SARSA with the SE kernel in (4). Using the product kernel in (5) leads to an approximately identical learning curve. The comparison for the policy generalization using different kernels, however, can also be achieved by a careful analysis in their associated reinforcement field plots.

The field plot helps in visualizing the control policy at arbitrary points in the state space, which can be inferred from the known particles retained in memory. From the model interpretability standpoint, a field plot makes it easier to conceptualize how the agent would behave according to its learned policy. Just like other physical fields such as an electric field, a physical quantity is defined everywhere in the space under influence. A group of charged particles create an electric force field, in which, if another charge particle is put in place somewhere in the field, it experiences an acceleration with an increasing kinetic energy along the direction of the net force exerted by these other particles. Similarly, in a reinforcement field, if the agent is placed at an arbitrary location, referred to as a test point, in the state space, it will move along a trajectory as governed by the combined influence from the experience

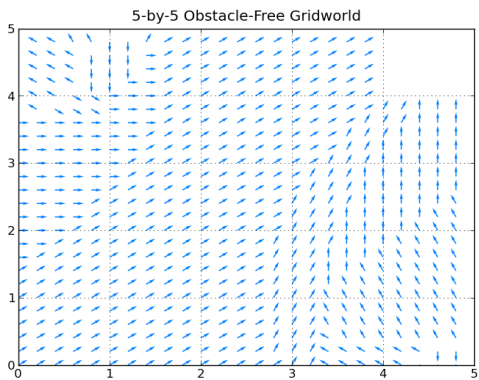

Fig. 8. Reinforcement field of a 5-by-5-partitioned grid world with no obstacles.

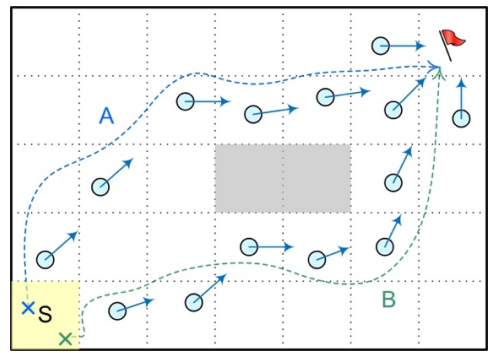

Fig. 9. A 7-by-5-partitioned grid world with areas filled with obstacles near the center zone. The dashed curve A and B represent two sample paths toward the goal that circumvent obstacles while minimizing path distances.

particles that established this field.

Fig. 8 illustrates the field plot induced by the noisy SE kernel in the 5-by-5(-partitioned) and obstacle-free grid world where the agent starts from an arbitrary location from within the lower left grid area and travels to the upper right grid representing the goal area, which is analogous to the setting in Fig. 4. The test points within each state partition are evenly spaced. Arrows in the field plot indicate the direction of motion reflecting action preferences at spatially different regions. For instance, the state partitions in the diagonal are in favor of (stochastic) actions 1 or 2 (see Fig. 1) that tends to resolve into the navigational direction aligned with the shortest path. Notice that both the uppermost and rightmost state partitions (especially those near the corners) reference some local control policies that are not consistent with the shortest path (in terms of misaligned arrows). This is resulted from the policy search converging prior to a thorough exploration of the state space. Indeed, this is desirable in that the agent is expected to identify an ideal control policy without the need to traverse the entire state space to shorten the learning time. To allow for a clean illustration, the size of the arrows is scaled according to the size of the grid and density of the test points, reflecting only the navigational directions rather than the actual step sizes. Due to the simplicity of this domain, the field plots induced by the noisy SE kernel and by the product kernel approximately follow the same pattern. Nonetheless, their policy generalization properties begin to exhibit subtle differences when obstacles are introduced to the grid world. This is illustrated in the following experiment where obstacles are placed along the shortest navigational paths toward the goal.

Specifically, imagine the scenario where the obstacle can cause a permanent damage to the agent and thus needs to be avoided by all means. This is in contrast to the state space boundary which merely forces the agent to navigate from within an allowable area (but without causing damage). Under this assumption, it is reasonable to assign a large negative reward (-100) whenever the agent hits an obstacle, whereas no penalty is given in addition to the per-step cost when the agent touches the border of the navigation area.

The environment setting that simulates the conditions mentioned above is depicted in terms of a 7-by-5(-partitioned)

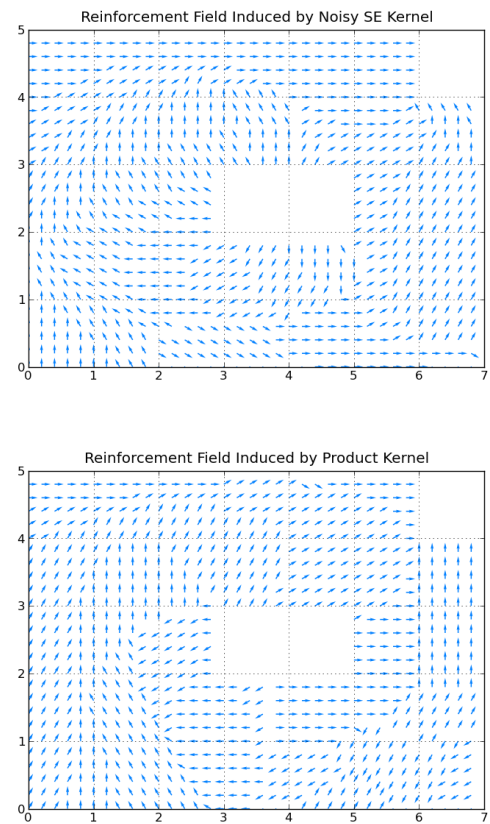

Fig. 10. Reinforcement fields in the 7-by-5-partitioned grid world with 2 center zones filled with obstacles. The top figure is the field plot associated with the noisy SE kernel while the bottom figure corresponds to the product kernel. The SE kernel-induced field exhibits relatively smoother streamlines, which tend to consistently guide the agent to steer away from the obstacles before merging back to the part of the field that (safely) drives the agent toward the goal.

grid world in Fig. 9, where obstacles are represented by two shaded state partitions (colored in gray) near the center area. The partition enclosing a start state ($S^{(0)}$) is marked by letter $S$ while the goal state is marked by a flag. Two approximately optimal paths toward the goal are depicted by the dashed curve A and B, respectively, both of which lead the agent to circumvent the region with obstacles while sharing approximately equal path lengths.

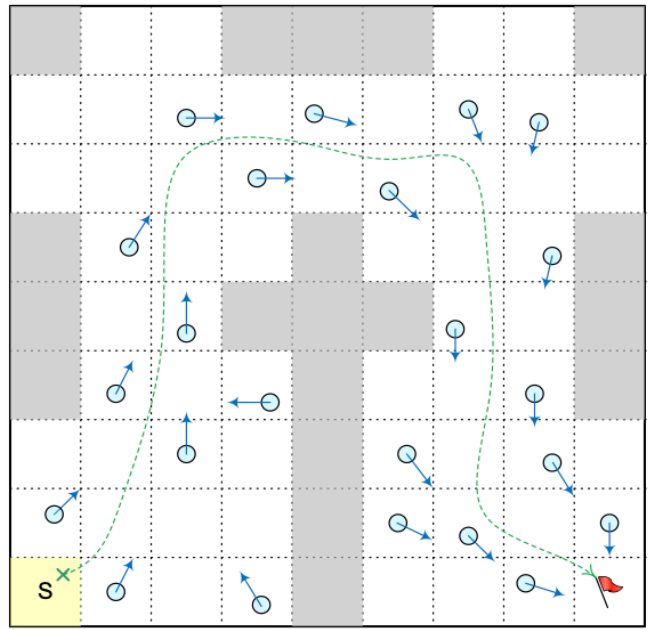

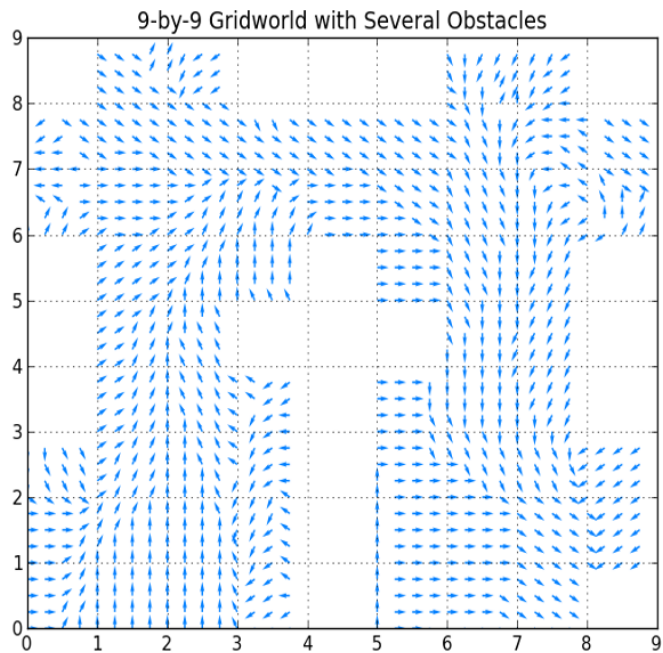

Fig. 11. The top figure is a 9-by-9-partitioned grid world with obstacles all over the state space. Like its 7-by-5 counterpart illustrated in Fig. 9, the green dashed curve, originating from the start state partition (denoted S), represents an (approximately) optimal path toward the goal state marked with a flag. Following the "streamlines" of the field, the agent is set on a path toward optimizing the global objective as it is guided by the distributed experience particles. The bottom figure is the corresponding reinforcement field plot, in which the obstacles along the way force the agent to take a detour to reach the goal. Most of the field turns out organized and structured, while a small part of the field may remain chaotic. In general, highly explored areas tend to exhibit a more consistently directed field whereas less explored areas can be relatively unpredictable.

Fig. 10 illustrates the corresponding reinforcement field established by the noisy SE kernel and the product kernel respectively. Interestingly, the product kernel exhibits a tendency for a greedy policy toward the shortest path in the proximity of the obstacle (under the same experimental settings including the number of training episodes). However, the policy induced by the SE kernel tends to be more conservative around the obstacles and thereby, provides a higher guarantee in escaping from the obstacle. For practical purposes, the ability to avoid obstacles may outweigh the benefit of following the shortest path in this scenario. Since actions are constantly influenced by random effects associated with possible errors in the actuator or unexpected environmental shifts, an "optimal action" does not necessarily resolve to an optimal real-time behavior. For instance, choosing the action that navigates to the east (i.e., action 3 in Fig. 1) along the southern border of the obstacle-filled region in Fig. 9 may not be desirable in practice even though a successful execution of such an action sequence may lead to a shorter path. Yet, the success rate for such an action sequence is in general very low unless the action outcome always meets the expectation.

A more complex scenario with a higher number of obstacle-filled regions is illustrated at the top of Fig. 11, in which the agent is required to take a significant detour in order to reach the goal area. The corresponding reinforcement field plot using a noisy SE kernel is given at the lower half of the figure. Like the cases we have seen earlier, the state space areas further away from the optimal path are likely to be more "distorted" due to the lack of exploration before RF-SARSA converges. Examples are the state space areas near the spatial coordinates (0.5,6.5), (0.5,7.5), and (8.5,7.5) in the field plot. Nevertheless, within these relatively unexplored areas, the agent still gave preference to escaping from the obstacle albeit not necessarily along an optimal path due to a lack of well-trained experience particles as supporting evidence.

Note that the standard SARSA can be considered as a special case of RF-SARSA by removing the parametric action model and the extra layer of fitness value estimate, thereby reducing the action operator to an ordinary action set with predefined behaviors. Nevertheless, doing so effectively drops the policy generalization capacity and often, more features need to be engineered into the state space to cope with potential complications from action dynamics, leading to an exponential growth in the state space. The need to increase the state feature due to the stochastic action dynamics will become apparent, for instance, in the task-assignment domain under time-varying computational resource to be illustrated in the next section.

### B. The Task-Scheduling Domain

To motivate a more complex parameterization of actions and demonstrate the G-SARSA using the CDLA framework introduced in Section V, we will examine an instance of the task-assignment domain briefly mentioned earlier in Section II.C. In particular, a task scheduling simulator was designed to simulate a continuous stream of user tasks to be assigned to a set of candidate servers of varying resource capacity (due to resource sharing among users). The specification of the experiment is given in Fig. 12. The goal is to determine an assignment policy that matches user tasks with compatible servers such that a given performance metric is optimized (e.g., minimum turnaround time). In this study, a match between a task and a server requires the *task type* to be identical to the *service type* of the server (see Fig. 12). This is to simulate the condition of the match-making process used in the HTCondor system [20], a workload management system that creates a high-throughput computing environment (HTC), in which the requirement statements of the task and those of the server must agree before a match can occur.

An example of a realistic requirement statement for a user task can involve the demand for a minimum CPU speed, memory size, or constraints on the platform as a computational environment. Similarly, a server may also have a requirement defined that only accepts certain job types and grants permissions to selected users. Under such match-making framework, a task will only execute on a server when both ends have compatible requirement definitions. For simplicity, the matching condition is simulated by checking the consistency between *task type* and *service type*, each of which is part of the variables that model a task and a server, respectively. If the matching condition holds, then the turnaround time is minimized when the target server has a high availability and relatively higher processing power, leading to a higher reward. Conversely, a task will fail to run under the condition of a mismatched requirement or insufficient resource capacity, all resulting in a negative reward. Note that a mutual agreement between task and server requirement still does not guarantee a success in terms of the task processing result. A deficient resource capacity relative to the task demand can also lead to a failed task, resulting in a negative reward identical to the mismatch case. This occurs, for instance, when the given task requires more memory than what is currently available in the matching resource.

| Parameter Spec. | Values |
|---|---|
| Task feature set (state) | ***Task type***, size, expected runtime |
| Server feature set (action) | ***Service type***, percentage CPU time, memory, disk space, CPU speed, job slots |
| Kernel *k* | *SE kernel + noise (see (4))* |
| Num. of Abstract Actions | *10 (assumed to be known)* |
| Model update cycle ***T*** | *10 state transitions towards 100* |

Fig. 12. Parameter specification for the task assignment domain.

Just like the navigation domain presented in the last section, the action parameters, which in this case characterize the server with time-varying resource capacity, are modeled by constrained random variables following designated probability distributions. In this study, the yield function mentioned in section II.B with a Gaussian random variation superimposed on the target value is used to simulate both the state and action variables.

An additional assumption is made to formulate an approximate match-making solution under MDP; that is, the state space variables, which collectively model the specification of the incoming tasks, are assumed to be static and fully observable once their values are determined (during the initialization phase in the simulation). That is, the task-specific feature values do not change over time and are fully accessible without added noise. In contrast, most of the server-specific variables (the set of action parameters) continue to vary after the assignment of their initial values except for one variable: *service type*. Since *service type* is used to simulate the requirement statement enforced by the machine owner, we assume that the statement does not vary over time as opposed to the other variables that characterize dynamically changing resource profiles. Doing so effectively simulates the condition that the resource capacity varies over time due to the sharing of multiple users across the network in addition to the varying demand of the incoming tasks. Moreover, the availability of the servers upon selection, which dictates the success rate of the task assignment to a target server, also varies according to a designated probability distribution. This is to simulate the condition where the server during the match-matching phase may not be always available due to limited job slots (as they are being occupied by other incoming tasks). Additionally, this condition is created to also illustrate the fluid behavior of actions, which can change over time: servers available in previous time steps may become unavailable in the future (either temporary or permanent, being occupied by other users). "Missing" actions could very well include the server previously thought to be the best assignment option (for certain tasks) and therefore must be substituted by other assignment options (i.e., different actions). This contrasts with standard RL where the action set is fixed and can never become inconsistent throughout the learning process.

With the standard RL formulation, the action set is determined by enumerating all possible assignments to available servers, which in general does not enable the agent to adapt the learned policy to the time-varying resource profile that characterizes the servers. Consequently, the standard SARSA is not an ideal policy learning algorithm for this domain. The key issue is that the same task assignment (by holding the task and server fixed) could have varying performance over time depending on the server availability and the remaining resource. In other words, the optimal policy identified earlier may not carry over into the future.

In the CDLA framework, on the other hand, the real-time server property is characterized by the parametric-action model, reflecting the task assignment action as a dynamic process contingent upon the various server attributes such as the CPU percentage, time, etc.

In this experiment, 6 action parameters are chosen to model the server, including CPU percentage, remaining memory, etc. (see Fig. 12). User tasks, on the other hand, are represented by 3 variables, including the matching criterion (i.e., *task type*) and 2 correlated variables: task size and expected runtime. Under these settings, each decision produces an augmented state with a 3-D state vector (task-specific) paired with a 6-D action vector (server-specific).

With the correlation assumption in the form of the kernel function in (4) – the SE kernel with weights associated with each feature dimension – each variable would contribute to the fitness value prediction differently. The task is thus to identify their relevance degree to the fitness value for a given state and action combination (as an assignment decision), where each unique "correlational pattern" effectively defines an abstract concept. For instance, identifying an abstract action representing a "matching concept" requires the agent to recognize the condition of *task type* and *server type* being equal (therefore matched) for a higher fitness value to occur.

Fig. 13 shows the experimental result comparing the G-SARSA against a random policy and a stripped-down version of the HTCondor's match-making algorithm, which always ensures a match prior to the task assignment. Fig. 14 illustrates an example of the abstract actions learned during the policy learning process. For instance, an abstract action that represents a *matching concept* is identified by the augmented state in which the state variable *task type* is identical to the action variable *service type* (the top 3 rows highlighted in blue). In contrast, abstract actions associated with non-match concepts at the bottom two rows (in yellow) lead to relatively lower fitness values and therefore are not preferable if actions that could lead to a match exist. Consequently, the most preferable strategy is for a task to always select an abstract action leading to matched server candidates and better yet, with the target server exhibiting a relatively higher resource capacity with high probability.

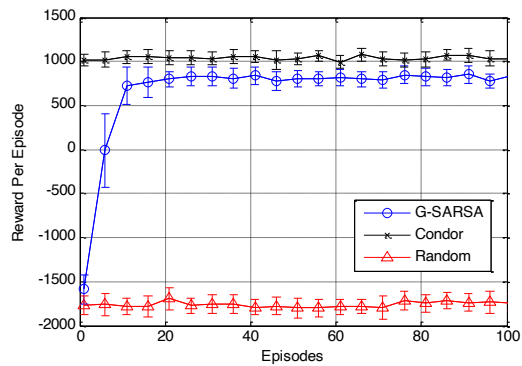

Fig. 13. Performance comparisons between the G-SARSA, a HTCondor-based match-making algorithm and a random policy. The softmax exploratory strategy used by the G-SARSA is eventually reduced to the ε-greedy strategy at low temperatures, where the reward per episode converges to slightly below the Condor benchmark.

|   | Task Type | Size | Expected Runtime | Service Type | %CPU Time | Fitness Value |
|---|-----------|------|------------------|--------------|-----------|---------------|
| 1 | 1 | 1.1 | 0.93 | 1 | 9.784 | 120.41 |
| 2 | 2 | 2.5 | 1.98 | 2 | 10.235 | 128.13 |
| 3 | 3 | 3.2 | 2.92 | 3 | 15.29 | 135.23 |
| 4 | 1 | 1.0 | 1.02 | 2 | 20.36 | -50.05 |
| 5 | 2 | 2.0 | 2.09 | 3 | 0.58 | -47.28 |

Fig. 14. Illustration of 5 different learned decision concepts where the top 3 rows in blue indicate success matches while the bottom 2 rows in yellow indicate failed matches. E.g., for tasks of type 1, the ideal policy (action) is the assignment to the servers of service type 1 (such that their indices agree) by selecting the blue abstract action (or concept), leading to a higher fitness value.

### C. Discussion

Recall from Section III.A that an action operator takes an input state, resolves stochastic effects inherent in the action, and then finally produces the next state. An action, in this general sense, effectively takes in the current state (a vector) and outputs its successor state (another vector), in which the vector-to-vector mapping is stochastic as governed by the action parameters as constrained random vector. Such action operator can be captured by a 2nd-order functional map $f_a: \mathbf{x}_a \to a(\mathbf{x}_a)$, where $a(\mathbf{x}_a): s \to s'$, where $\mathbf{x}_a$ in this case denotes a random vector of action parameters. Depending on the task domain, such action parameter-induced mapping can be explicit or implicit. In the navigation domain (Section A), for instance, the action operator $f_a$ can be expressed in an analytical form as a parameterized matrix multiplication (see (11)); in contrast, the action operator in the task-assignment domain is only implicitly defined. Since the action in the navigation domain has a direct relation to the next state (which depends on the current position and the direction along which the agent chooses to move), it is straightforward to specify the action operator in the form of a matrix. This is true in general for a linear dynamical system in which the time-varying output can modeled by a linear transformation on an input.

However, within the task scheduler that assigns an ongoing stream of tasks to a pool of dynamically varying servers (in terms of resource profile and availability), the scheduled task, as the current state, does not necessarily have a directly identifiable relationship with the next task, as the successor state, in a manner that depends on the chosen server, which corresponds to a task-assignment action. For one thing, a candidate server is often shared by multiple users across the network and possibly without affiliation to each other. For another, the user does not have the access to the knowledge of the server availability in a timely fashion to schedule his or her tasks accordingly. As a result, the correlation within the task stream is often (nearly) independent of the target server albeit the possibility for any two or more tasks to assume interdependency. In such case, the action operator $f_a$ takes on a family of functions of the form: $a(\cdot): \mathbb{R}^m \to \mathbb{R}$, where $a(\cdot)$ maps from a state (say, of dimension $m$) to a real value (e.g., a reward). In other words, the decision processes underlying the task-assignment domain considers both the context (the state), the action and their combined effect but ignores the notion of successor states and delayed reward. This is referred to as the contextual bandit problem, or one-step RL, a special case of a full RL problem, since the optimal action depends solely on the current state but not on the successor states.

For simplicity, the empirical study in Section B assumes that the successor task is independent of the server to which the current task is assigned. Consequently, it is unnecessary to explicitly specify the action operator that relates the current task to the next one other than specifying the action parameters collectively as a random process. Nonetheless, the notion of action operator still applies in the task-assignment domain since the augmented state, as the byproduct of an action operator, is produced in the same manner as in the navigation domain through the following logical sequence: i) an input state ($s \in S$) is made available by the scheduler as an outgoing task awaiting for a matching server ii) the contextual bandit-style action operator, $a(\mathbf{x}_a): s \to \mathbb{R}$, takes on this input task (characterized by the state $s \in S$), resolves the uncertainty from the action by taking a snapshot of the resource capacity of the chosen server $\mathbf{x}_a$ upon the assignment of the task, and gets a scaler feedback (i.e. a reward $r \in \mathbb{R}$) iii) the next task again is made available as the next state ($s' \in S$), although this time, the new task does not have a direct relationship with the predecessor task ($s \in S$) through the action taken previously. Perhaps a natural way to model the task assignment problem is to use a generalized form of contextual bandits [36], which, as illustrated, can be formulated as a special case of the GRL.

## VII. CONCLUSION AND FUTURE WORK

This paper generalizes the standard RL through a two-stage construction. The first stage introduces the notion of the reinforcement field that serves as a policy generalization mechanism built on top of the RKHS. Essentially, the reinforcement-field construct is a physics-inspired, kernel-based, sample-driven representation that is rendered possible through extending the action formulation by moving from the primitive action in standard RL to the parametric-action model and the action operator. In this new action-oriented perspective for policy learning, the action set is no longer being merely interpreted as a pre-defined, fixed set of decision choices (whether they be discrete or continuous) but instead, collectively assumes a more flexible representation as an action operator, parameterized by an arbitrary number of (constrained) random variables as defined by the task.

The action operator defines the state transition operationally such that the action can be considered as the property of an "external field" that acts upon an input state and subsequently yields another state, following the rule as defined by the field and aligned with given optimization objective such as maximizing the accumulated reward. In this manner, various processes that underly the policy search can be integrated as a unified continuous process, in which state transitions are induced by local constraints enforced by the action operator, step by step, such that the resulting state trajectory leads to a maximized accumulated reward as the global objective. The intuition behind such formulation originally emerged from the observations of physical fields and the Lagragian mechanics (e.g., how particles move in an electromagnetic field), which are briefly mentioned in Section III.A.

The second stage builds on top of the reinforcement field and subsequently formulates the action-oriented policy abstraction to reduce decision points per state by reusing the very same kernel function from the GPR as the cluster assumption for grouping similar decisions, taking into account all variables within the state and the action along with their corresponding fitness value (representing the "fitness" of their binding as two interactive entities). Pattern discovery from within each cluster-based action abstraction (comprising similar decision points) can now serve as a basis for identifying the conditions by which compatible bindings between states and actions can be established – where they play the role of two mutually interactive entities like the tasks as states and the assignments to servers as actions – while keeping the global objective of reaching a high long-term payoff (e.g., minimizing the error rate resulted from the task assignment). The notion of abstract actions effectively provides an instrument for learning a context-dependent control policy such as automatically learning the match-making criteria for compatible task-and-server assignments as illustrated in Section VI.B.

### A. The Learning Architecture Perspective

The interleaving process between the kernel-based value function approximator like GPR and the kernel-based decision concept formulation leads to the concept-driven learning architecture (CDLA) that underlies the GRL framework. All the components in the CDLA are represented directly or indirectly through the kernel function that serves as the fundamental building block for all learning processes including value prediction, policy search, working memory update, and decision concept formation. Under the CDLA, the control policy can be learned at the scale of primitive actions, parametric actions, and abstract actions.

Specifically, the GRL under the CDLA consists of the following primary components:

- The base RL layer, which runs a selected RL algorithm under MDP (e.g., SARSA).
- Parametric action model and action operator, which transform the primitive action into a parameterized functional form that acts on the state in a systematic way while factoring in uncertainty in action behaviors.
- The working memory coupled with a TD-based memory update mechanism (particle reinforcement), which references a set of progressively updated experience particles as a functional data set that establishes a reinforcement field (through the kernel) to enable policy inference and generalization, and structurally forms a graph (again through the kernel) on which decision concepts can be learned.
- The GPR layer, which plays the role of an adaptive, sample-driven (fitness) value function approximator that receives training signals from the base layer while enabling the value prediction over the augmented state representation; automatic similarity metric learning occurs through periodical kernel hyperparameter tuning via the ARD procedure.
- Abstract decision concept learner, which treats the working memory as a graph, on which a selected clustering algorithm is applied to identify contextually similar decisions, based on which abstract actions are formulated; the policy learning on the level of abstract actions is made possible via memory association, a self-organizing property rendered by the kernel function from the GPR layer.

Summarized in Fig. 15 in terms of a robot-like workflow, the GRL framework under the CDLA consists of kernelized components that are marked by a circled "k" in yellow as well as indirectly-kernelized components that are marked by a green circled "k". The two types of policy functions, one being at the scale of the parametric actions and the other at abstract actions (depicted by the left and the right "arm"), are indirectly kernelized due to their dependency on the GPR as the kernel-based functional approximator.

### B. Linking Policy Search to Metric Learning

We use the terminology of "kernelization" to describe the process of endowing the algorithmic components with controllable behaviors regulated by the underlying kernel function. The kernel function – or broadly speaking, as we envisioned, any metric learning system – will play an integral part in extending the GRL framework further, since, to implement a next-generation RL agent that links together adaptive action modeling, policy search, and decision concept formulation, among other components, the learning system would benefit greatly from the associative-memory capacity as part of the feedback loop. The idea is that, as the agent learns to make optimal decisions, it also attempts to improve upon its world model by drawing connections from within the historical decisions made in the past; an improved world model subsequently prompts greater decision-making ability, and the iterative process continues on. With this decision-specific cognitive model, the agent can learn to reuse "good" decisions made from similar situations in the past as it attempts to fulfill a global optimization objective while moderately "improvising" via the randomness inherent in its action operator. On the contrary, "bad" decisions can serve as counterexamples to be avoided or as constraints that steer the action ever so slightly away from the original action parameter settings that had led to poor results.

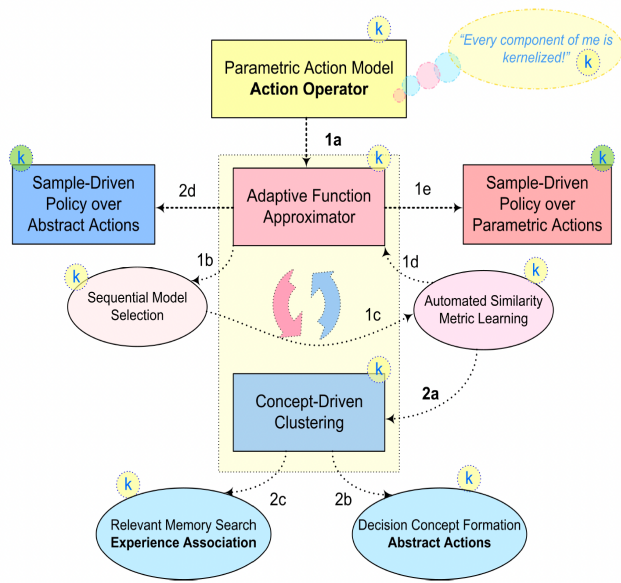

Fig. 15. The GRL framework under CDLA depicted as a workflow shaped like a robot. The periodic iterative process between the fitness value prediction (upper torso) and the decision concept discovery (lower torso) constitutes the heartbeat of a GRL agent. The conceptual sequence from 1a to 1e corresponds to the workflow that progressively updates the reinforcement field as a policy generalization construct. The sequence from 2a to 2d, on the other hand, represents the workflow towards establishing an abstract control policy based on a set of coherent decision concepts as high-level control decisions.

In GRL, the story unfolds from the parametric action model: as the agent explores the environment, each decision it makes along the way is captured by an experience particle kept in its working memory. The collection of particles implicitly defines a graph, in which particles, comprising snapshots of decisions and their fitness measures, correspond to the vertices, while the edges are implicitly formed via the particle similarity as measured by the kernel function.

Representing experience particles – a 3-tuple comprising the state vector, the action vector and its functional response –

through the kernel function, effectively leads to a policy-generalization construct, reinforcement field, where at any point in the state space, there is a corresponding action configuration (or configurations) that a particle occupying that state must take in order to maximize a global objective such as maximal utility. The optimal binding between the state and the action thus constitutes a localized policy with a measurable quantity referred to as the fitness value defined at every point in the reinforcement field. The fitness value is measured through the GPR layer that inherently depends on the kernel function such that the value of any new particle (factoring in both state and action configurations) is dominated by the known surrounding particles (e.g., particles in the nearby state space, or within the same state partition) with relatively higher degrees of correlation as compared to those from distant particles. The reinforcement field also inherits properties from the RKHS such that small changes in the state and action configurations would result in correlated fitness values; that is, the state-action combinations in the neighboring (augmented) state space are expected to share similar fitness values. This property also dictates how the agent evaluates new decision hypotheses: recall from Section IV.A that the agent infers the best action at an unexplored query state by forming hypothetical states and running experience association with the past memory, which is also driven by the kernel function as a correlation hypothesis.

The connection between the control policy and the kernel function can be visualized in terms of a field plot, in which each position represents a state with a corresponding arrow pointing in the direction consistent with the control policy. These "pointers" can be visualized when the action parameters as random vectors live in lower dimensional spaces such as the 2D field plots in Fig. 10. If a state partition (with its granularity as a tunable variable) is sufficiently explored such that an optimal policy can be inferred from the trade-off between particles of both polarities (negatives and positives), then the arrow will in principle point in the direction that fulfills the global objective. Conceptually, the arrow in the field represents a localized control policy that also generalizes into the neighboring state space by virtue of the experience association (see also Section VI.A). Incidentally, as a design choice for the GRL framework, one could also seek to generalize the binary-polarity model in the particle reinforcement, covered in Section IV.A, into a "colored version," where each (primitive) action, for instance, is associated with its own fitness function, maintaining action-specific "charged" particles with positive TDs or negative TDs.

The ability to relate from one experience particle to the others leads to an instance of abstract action model with a decision concept hierarchy under CDLA (Section V), where the agent learns to form higher-level abstractions over decision contexts. Under CDLA, a control policy can be defined on the level of abstract actions, or decision concepts, on top of the parametric action model and the primitive action set. Experience association again plays an important role in that a decision concept itself is not directly executable in the state space, but needs to be reduced to concrete, primitive actions in order for the state transition to occur. The process of inferring the best action(s) from an abstract concept (referred to as action resolution) is accomplished by identifying the most relevant pieces of memory from a group of correlated particles (although we have only demonstrated a greedy approach in this paper).

With a properly parameterized kernel function, each abstract action, essentially a cluster of particles, can also be shown to reveal correlational structures across variables that define the state and the action. As an example, we used a specialized SE kernel with each state and action variable weighted by a hyperparameter in their joint vector representation (although in principle, it is possible to learn a full covariance matrix). Subsequently, through the marginal likelihood maximization via ARD, the optimal configuration of these hyperparameters can then be used to identify, for each abstract action, the conditions to hold for attaining high fitness values in terms of (interpretable) relationships within the state and the action variables. A subset of abstract actions can correspond to, for instance, "positive" decision concepts comprising historical decision points (i.e., experience particles in memory) that reference relatively high fitness values, reflecting beneficial bindings between constituent states, or contexts, and the actions taken within these contexts. Similarly, one can also identify "negative" concepts represented by clusters with decision points referencing relatively low fitness values. To this end, we showed in a simulated task-assignment domain that CDLA can effectively identify the requirement statement necessary to optimize the final task-assignment policy (Section VI.B).

### C. Future Work

The algorithmic components proposed in this paper, such as SARSA as the base layer and the GPR as the fitness function approximator, are meant to be an instance of how the GRL framework can be implemented, where the "L" really focuses on the adaptive policy learning processes for complex time-varying systems based on an alternative view of action formulation (the "G") – the parametric action model, the action operator, and the abstract action. Components in the learning architecture, CDLA, are linked together through the kernel machinery in this paper but other metric learning methods that are tunable as new experience particles are collected can be applicable to GRL.

The future work will involve further assessments of the GRL framework developed in this paper in various application domains. The extension of the GRL in POMDPs and in probabilistic experience association, as mentioned in Section IV.C, are of interest for increasing the adaptability of GRL in complex, time-varying systems. Since every component in the CDLA can be kernelized (or in general formulated as a metric learning process), dedicating an extra layer specifically for the representation learning of decision similarity can potentially improve the performance of the GRL on multiple fronts through the interconnected processes inherent in the CDLA: value function approximation, policy search, policy generalization in the reinforcement field, experience association (or other forms of associative memory), particle reinforcement (or other self-organizing properties for the working memory), and decision concept formulation with the control policy learned on multiple levels of abstractions.

From the kernel machine's perspective, the inputs to kernel functions are certainly not limited to the vector form but can also include more complex, symbolic objects including sets, graphs, strings, tensors, etc., as long as they correspond to valid inner products in the feature space [1], [15]. For instance, graph representation can potentially be used to encode more complex interactions between the state and the action as two decision-specific constructs. From the perspective of metric learning and representation learning, it is of interest to find other richer expressions for the binding between the state and the action, representing localized policies that also generalize to the neighboring state space. Subsequently, the resulting decision metric can be used as the correlation hypothesis using the very same learning architecture, CDLA, for policy learning and inference at various levels of action abstractions, from the primitive, to the parametric, and the abstract.